\documentclass[lettersize,journal]{IEEEtran}
\usepackage{amsmath,amsfonts}
\usepackage{algorithmic}
\usepackage{algorithm}
\usepackage{array}
\usepackage[caption=false,font=normalsize,labelfont=sf,textfont=sf]{subfig}
\usepackage{textcomp}
\usepackage{stfloats}
\usepackage{url}
\usepackage{verbatim}
\usepackage{graphicx}
\usepackage{cite}
\usepackage{soul}
\usepackage{xcolor}
\usepackage{fancyhdr}

\begin{document}

\title{Integrating Specialized and Generic Agent Motion Prediction with Dynamic Occupancy Grid Maps}

\author{Rabbia Asghar, Lukas Rummelhard, Wenqian Liu, Anne Spalanzani, Christian Laugier\\
}

\renewcommand\fbox{\fcolorbox{red}{white}}
\setlength{\fboxrule}{2pt} 

\markboth{Journal of xx,~Vol.~xx, No.~x, Month~xx}%
{Shell \MakeLowercase{\textit{et al.}}: A Sample Article Using IEEEtran.cls for IEEE Journals}

\fancypagestyle{firstpage}{
  \fancyhf{}
  \renewcommand{\headrulewidth}{0pt}
  \fancyfoot[C]{\small This work has been submitted to the IEEE for possible publication. Copyright may be transferred without notice, after which this version may no longer be accessible.}
}
\maketitle
\thispagestyle{firstpage}
\begin{abstract}
Accurate prediction of driving scene is a challenging task due to uncertainty in sensor data, the complex behaviors of agents, and the possibility of multiple feasible futures. Existing prediction methods using occupancy grid maps primarily focus on agent-agnostic scene predictions, while agent-specific predictions provide specialized behavior insights with the help of semantic information. However, both paradigms face distinct limitations: agent-agnostic models struggle to capture the behavioral complexities of dynamic actors, whereas agent-specific approaches fail to generalize to poorly perceived or unrecognized agents; combining both enables robust and safer motion forecasting. To address this, we propose a unified framework by leveraging Dynamic Occupancy Grid Maps within a streamlined temporal decoding pipeline to simultaneously predict future occupancy state grids, vehicle grids, and scene flow grids. Relying on a lightweight spatiotemporal backbone, our approach is centered on a tailored, interdependent loss function that captures inter-grid dependencies and enables diverse future predictions. By using occupancy state information to enforce flow-guided transitions, the loss function acts as a regularizer that directs occupancy evolution while accounting for obstacles and occlusions. Consequently, the model not only predicts the specific behaviors of vehicle agents, but also identifies other dynamic entities and anticipates their evolution within the complex scene. Evaluations on real-world nuScenes and Woven Planet datasets demonstrate superior prediction performances for dynamic vehicles and generic dynamic scene elements compared to baseline methods.
\end{abstract}


\section{INTRODUCTION} \label{sec:introduction}
\IEEEPARstart{S}{cene} understanding is a fundamental challenge in autonomous driving, where accurate perception and prediction of the environment are essential for safety. This challenge arises from various factors such as partial or unreliable sensing, unfamiliar environments, complex behavior of agents, and the possibility of diverse future behaviors \cite{yurtsever2020survey}. Given these complexities, most methods tackle the problem using a modular approach by detecting and tracking agents and predict their specific behavior, while some focus on agent-agnostic occupancy and motion estimation.

To effectively predict motion, both agent-specific and agent-agnostic motion predictions are important in their respective contexts. Understanding the behaviors of agents in relation to their surroundings allows for better anticipation of motion and interactions, enabling proactive planning based on observed behaviors \cite{mozaffari2020deep}. Meanwhile, generalized motion prediction aids in recognizing general motion patterns, which can be essential for responding to unforeseen situations, or when encountering unfamiliar or poorly recognized agents. Together, these predictions enhance overall safety by avoiding reliance solely on assumptions about recognized agents' behavior.

Commonly, dynamic occupancy grid maps (DOGMs) are used to predict generic static and dynamic elements within the environment in a bird's-eye view (BEV) grid. Probabilistic DOGMs \cite{lukas22} enable the modeling of sensor and prediction uncertainties, offering a significant advantage over traditional object detection-and-tracking methods that employ confidence thresholding. Such frameworks facilitate dense probabilistic occupancy representation, {enable velocity estimation of dynamic grid cells, and represent uncertainty in partially or fully occluded regions as unknown occupancy state}, all while remaining agent-agnostic.

\begin{figure}[t]
	\centering
	    \includegraphics[trim={0 0 0 0},clip,width=0.8\columnwidth]{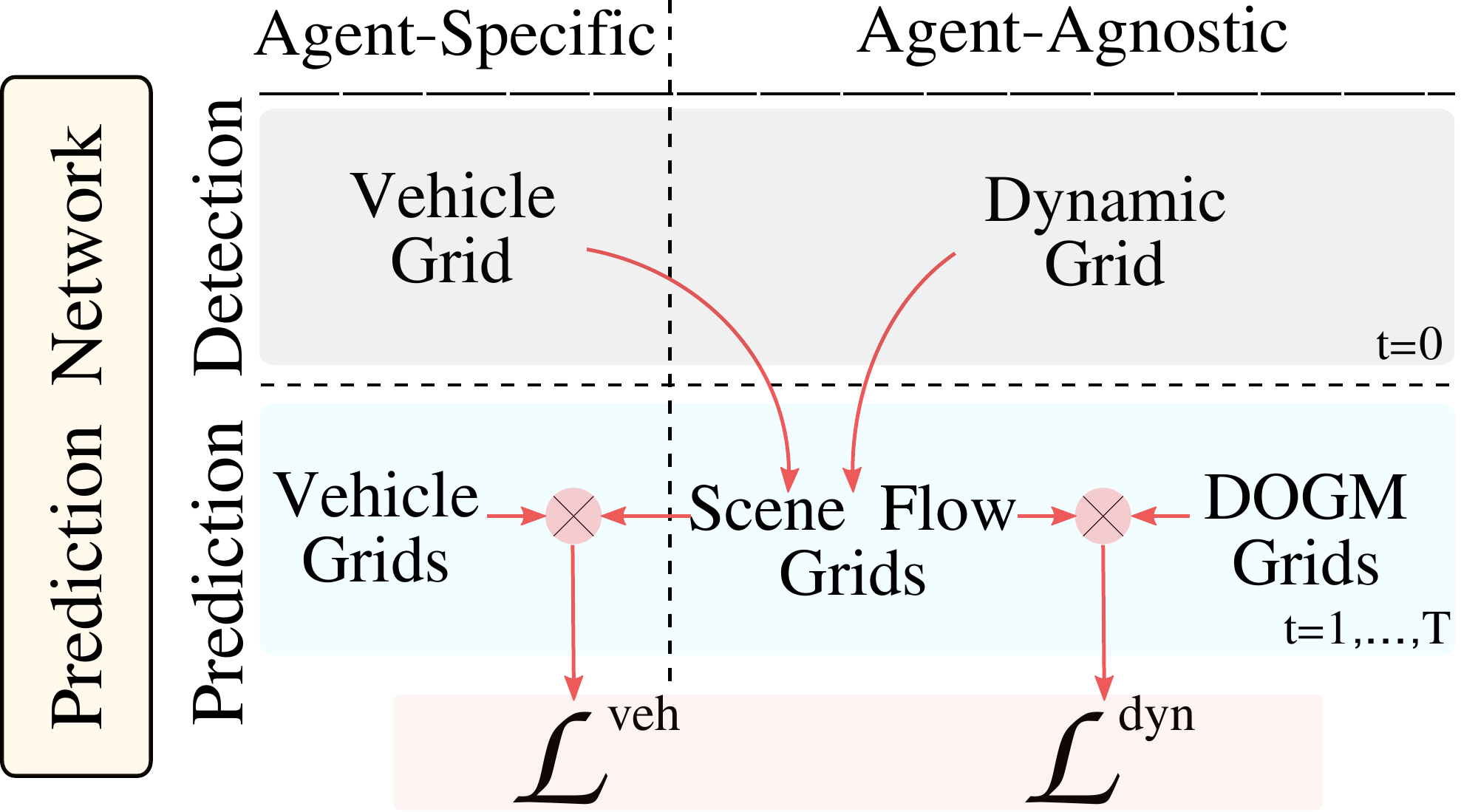}
\caption{\small Our network generates detection and prediction grids, for agent-specific and agent-agnostic behaviors. Scene flow grids play a central role in bridging these two categories, leveraging flow-guided occupancy predictions and optimizing them through tailored loss functions.}
\label{fig:intro}
\end{figure}
However, DOGMs come with their own share of limitations. {Since they do not leverage learned shape or semantic priors, }DOGMs may fail to capture the intricate shapes of agents, such as large vehicles \cite{asghar2022allo}. In scenarios where sensor information is limited, these grids may further struggle to assign reliable occupancy probabilities and to effectively distinguishing between static and dynamic components in the environment. {Integrating deep learning techniques with agent semantic information offers a promising direction to mitigate these limitations \cite{asghar2023vehicle,toyungyernsub2024predicting}, while enabling a unified framework that combines agent-agnostic and agent-specific motion prediction.} 
Furthermore, our scene flow grids together with flow-guided occupancy predictions enable the framework to bridge both agent-specific and generic agent motion prediction, Fig. \ref{fig:intro}, and enhance the network ability to learn diverse future predictions.

In this work, we explore scene evolution prediction by combining probabilistic dynamic occupancy grid maps with vehicle segmentation grids.
We employ a lightweight and established pipeline to detect agents and identify dynamic actors, while simultaneously predicting multi-task future grids.
Our contributions are summarized as follows: i) {Multi-head holistic representation:} a novel approach that combines the agent-specific behavior prediction and agent-agnostic motion prediction through a multi-head decoding structure that simultaneously generates vehicle grids, probabilistic occupancy grids, and flow grids to provide a holistic representation of scene evolution; ii) {Consistent multi-task optimization:} a set of interdependent optimization objectives that explicitly link decoder outputs by utilizing scene flow as a supervisory bridge. By leveraging available occupancy states to enforce motion-occupancy consistency,  the loss function enables the framework to learn and predict the motion of any dynamic entity in the scene while capturing diverse and plausible future behaviors. Experimental results on two real-world datasets demonstrate that our method generalizes well across diverse environments, consistently showing strong performance in dynamic scene prediction.
\section{RELATED WORK} \label{sec:related-work}

\subsubsection{Occupancy Grid Map (OGM) Predictions}
In the state-of-the-art literature, the spatio-temporal problem of predicting ego-centric future OGMs is often treated as a video prediction task, employing deep-learning methods involving combinations of convolutional neural networks (CNNs) and recurrent neural networks (RNNs) \cite{jeon2018traffic, dequaire2017deep, mohajerin2019multi, itkina2019dynamic}.
These frameworks process raw sensor data and utilize ego-vehicle localization \cite{reid2019localization} to generate and predict grids typically without the need for manual ground-truth annotations.
Common challenges that persist across these prediction approaches include blurriness, loss of scene structure and the disappearance of dynamic agents. Addressing these problems, studies by \cite{Toyungyernsub2022, schreiber2019long} predict the static and dynamic scenes separately while \cite{lange2021attention} proposed using attention-augmented ConvLSTM (Convolutional Long Short Term Memory) to make long-term predictions. Different from these, we \cite{asghar2022allo} explored future DOGM prediction within a fixed reference frame, preserving scene integrity, particularly during ego-vehicle maneuvers.

\subsubsection{Agent-specific predictions}
Prevalent state-of-the-art work in autonomous driving literature relies on heavily processed data and the availability of annotated agents for motion forecasting, utilizing deep learning architectures such as Graph Neural 
Networks \cite{gao2020vectornet}, LSTMs \cite{salzmann2020trajectron++}, and attention mechanisms \cite{varadarajan2022multipath++, schafer2022context}. 
These methods exhibit varying dependencies on HD maps \cite{diaz2022hd} and 
goal-oriented predictions, with some approaches operating in map-free settings 
by focusing on social interactions \cite{schmidt2022crat, benrachou2023improving}, while others exploit vectorized lane representations \cite{gomez2023improving, gomez2023efficient}.
Meanwhile, many end-to-end methods have emerged to address the fusion of detection, tracking and prediction modules \cite{liang2020pnpnet, wu2020motionnet, luo2018fast}. MotionNet  \cite{wu2020motionnet} introduced a joint perception and motion prediction network that outputs BEV maps, with each grid cell encoding object category and motion information. In relation to occupancy grid use, Fiery \cite{fiery2021}, MP3 \cite{casas2021mp3} and FishingNet \cite{hendy2020fishing} predict future occupancy grids with semantic labels for varying agents such as vehicles, pedestrians, cyclists. Occupancy flow fields \cite{mahjourian2022occupancy} addressed motion forecasting as occupancy and flow grid prediction specific to the vehicles and pedestrians in the scene. We extend the representation proposed in this method by applying it to noisy input and to both agent-specific and agent-agnostic dynamic occupancy predictions.

\subsubsection{Combined Agent-agnostic and Agent-specific predictions}
Recently, some approaches have combined agent semantic information with DOGMs to enhance prediction capabilities. \cite{mann2022predicting} relied on vehicle masks and used thresholding to convert the OGMs into binary grids while \cite{toyungyernsub2024predicting} incorporated agent and environment semantics with OGMs to predict future occupancy grids without semantics. We \cite{asghar2023vehicle, asghar2024flow} explored integrating vehicle semantics with DOGMs with map and occupancy flow configurations respectively, but only predicted the future vehicle and flow grids. 

Significant challenges remain in achieving comprehensive situational awareness that balances agent-specific prediction with a robust account of sensor uncertainties and unclassified objects in unforeseen environments.
To address this, we jointly forecast vehicle grids, scene-wide flow, and future DOGMs. Our multi-task approach leverages semantic priors to improve vehicle behavior prediction while utilizing DOGM occupancy states to address the uncertainty propagation inherent in environments with occlusions or diverse, non-vehicle agents.
\section{PROBLEM FORMULATION} \label{sec:prob-formulation}
\begin{figure*}[ht]
	\centering
	    \includegraphics[trim={0 0 0 0},clip,width=1.5\columnwidth]{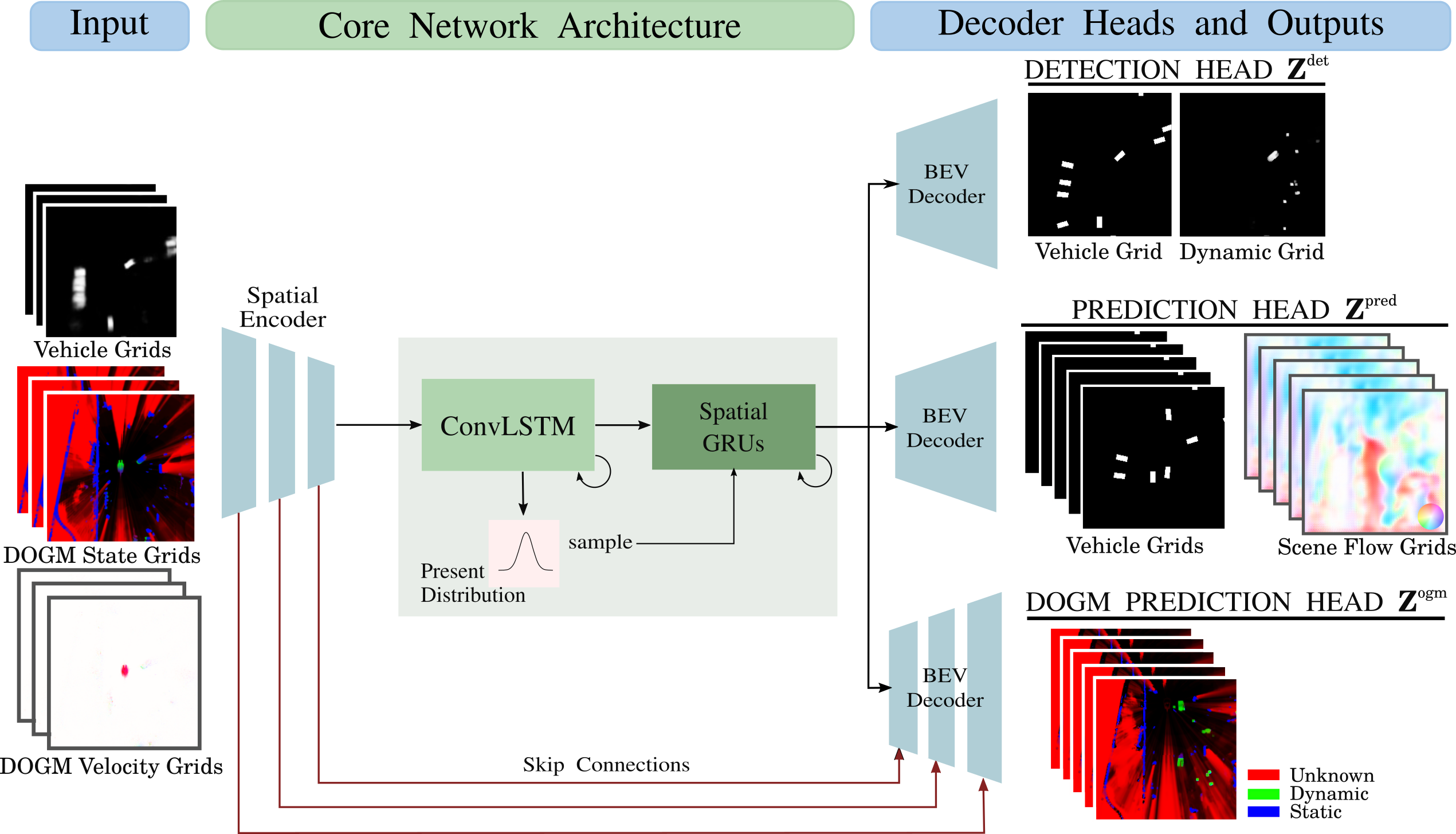}
\caption{\small Proposed prediction framework: Input sequence of vehicle grids, DOGM occupancy state grids and associated velocity grids capture the past scene evolution. The network takes inspiration from video prediction, conditional variational approach, skip connection architecture and predicts i) vehicle and dynamic grid at current timestep in the detection head $\mathbf{Z}^{\text{det}}$, ii) sequence of future vehicle and scene flow grids in the prediction $\mathbf{Z}_{1:T}^{\text{pred}}$ and iii) sequence of three-channel occupancy state grids in the DOGM prediction head $\mathbf{Z}_{1:T}^{\text{ogm}}$.}
\label{fig:overview}
\end{figure*}
We formally define the task of motion prediction in driving scene, using a BEV grid representation, Fig. \ref{fig:overview}.
Let the input sequence be denoted by $\mathbf{X}_{t-N:t}$, where each frame $\mathbf{X}_i \in \mathbb{R}^{6 \times w \times h}$ represents the environment state, $w$ the width and $h$ the height of the grid. $\mathbf{X}_i$ consists of three probabilistic occupancy states, two velocity components, and one vehicle segmentation mask.

Given the past $N$ time steps, we formulate the task as a supervised learning problem, seeking a neural network $f_{\theta} \colon \mathbf{X} \to \mathbf{\hat{Y}}$ that maps the history $\mathbf{X}_{t-N:t}$ to a set of multi-head future predictions $\mathbf{\hat{Y}}$ over a horizon of $T$ steps:

\begin{equation*}
    \mathbf{\hat{Y}} = \{ \mathbf{Z}^{\text{det}}, \mathbf{Z}_{1:T}^{\text{pred}}, \mathbf{Z}_{1:T}^{\text{ogm}} \}
\end{equation*}
where the output heads are defined as follows:\\
\textbf{i) Detection Head} ($\mathbf{Z}^{\text{det}} \in \mathbb{R}^{2 \times w \times h}$): Predicts the vehicle grid $\mathbf{D}_{\text{det}}^{\text{veh}}$ and agent-agnostic dynamic occupancy grid $\mathbf{D}_{\text{det}}^{\text{dyn}}$ at the current time step $t$, \\
\textbf{ii) Prediction Head} ($\mathbf{Z}_{1:T}^{\text{pred}} \in \mathbb{R}^{T \times 3 \times w \times h}$): Forecasts the future vehicle grids and scene flow fields $\{ \mathbf{P}_{\tau}^{\text{veh}}, \mathbf{P}_{\tau}^{\text{flow}} \}_{\tau=1}^{T}$,\\
\textbf{iii) DOGM Prediction Head} ($\mathbf{Z}_{1:T}^{\text{ogm}} \in \mathbb{R}^{T \times 3 \times w \times h}$): Forecasts the sequence of future agent-agnostic probabilistic occupancy states $\{ \mathbf{O}_{\tau}^{\text{unk}}, \mathbf{O}_{\tau}^{\text{stat}}, \mathbf{O}_{\tau}^{\text{dyn}} \}_{\tau=1}^{T}$ representing the unknown, static, and dynamic occupancy channels respectively.

The optimal parameters $\theta^*$ are obtained by minimizing a composite loss function $\mathcal{L}$ against the ground truth $\mathbf{Y}$: 
\begin{equation} 
\theta^* = \arg \min_\theta \mathcal{L}(\mathbf{\hat{Y}}, \mathbf{Y}) 
\end{equation}
\section{SYSTEM OVERVIEW} \label{sec:sys-overview}
The architectural framework of our model is modular, allowing for interchangeable spatio-temporal backbones and input configurations while anchoring the predictive performance within a shared cross-task optimization objective. We first define the multi-channel input modalities, followed by the core network architecture and multi-head decoder. We then explain our main contribution with the flow-guided occupancy prediction mechanism and the tailored loss functions.
\subsection{Input Modalities: DOGM and Inclusion of Semantics}
The input to our model is a sequence of multi-channel BEV tensors  $\mathbf{X}_{t-N:t}$, constructed by the concatenation of the DOGM and the vehicle segmentation grid.

\subsubsection{Dynamic occupancy grid maps} DOGMs provide a BEV grid-based representation of the environment. Each cell within the grid is independent and is estimated in parallel by the system, offering information about its occupancy state and associated dynamics in an agent-agnostic manner.
We utilize a Bayesian dynamic occupancy grid filter \cite{lukas22}, to process raw LiDAR data and estimate the probabilities for 4 possible occupancy states in each grid cell: i) free, ii) occupied and static, iii) occupied and dynamic, and iv) unknown.
Fig. \ref{fig:input} represents these occupancy states.
Note that the unknown occupancy state represent regions with insufficient sensor data, where the true occupancy state may be free or occupied. Additional to the occupancy state grids, the framework computes velocity estimates for each grid cell, that are represented by mixture of grids and particles. 

In our proposed model, the input $\mathbf{X}_t \in \mathbb{R}^{6 \times w \times h}$ comprises six channels, five of which are obtained from this DOGM framework. Among these five channels, three represent the occupancy states $\{\mathbf{O}^{\text{unk}}, \mathbf{O}^{\text{stat}}, \mathbf{O}^{\text{dyn}}\}$ and two represent the velocity channels $\mathbf{V} = \{\mathbf{v}_x, \mathbf{v}_y\}$. Three occupancy states are represented as an RGB image in this work, with red, green and blue channels representing unknown state, dynamic occupied and static occupied state respectively. These images implicitly  represent free state within the grid as black areas. 

\subsubsection{Semantic Segmentation}\label{subsec:semantics}
\begin{figure}[t]
	\centering
	    \includegraphics[trim={0 0 0 0},clip,width=0.97\columnwidth]{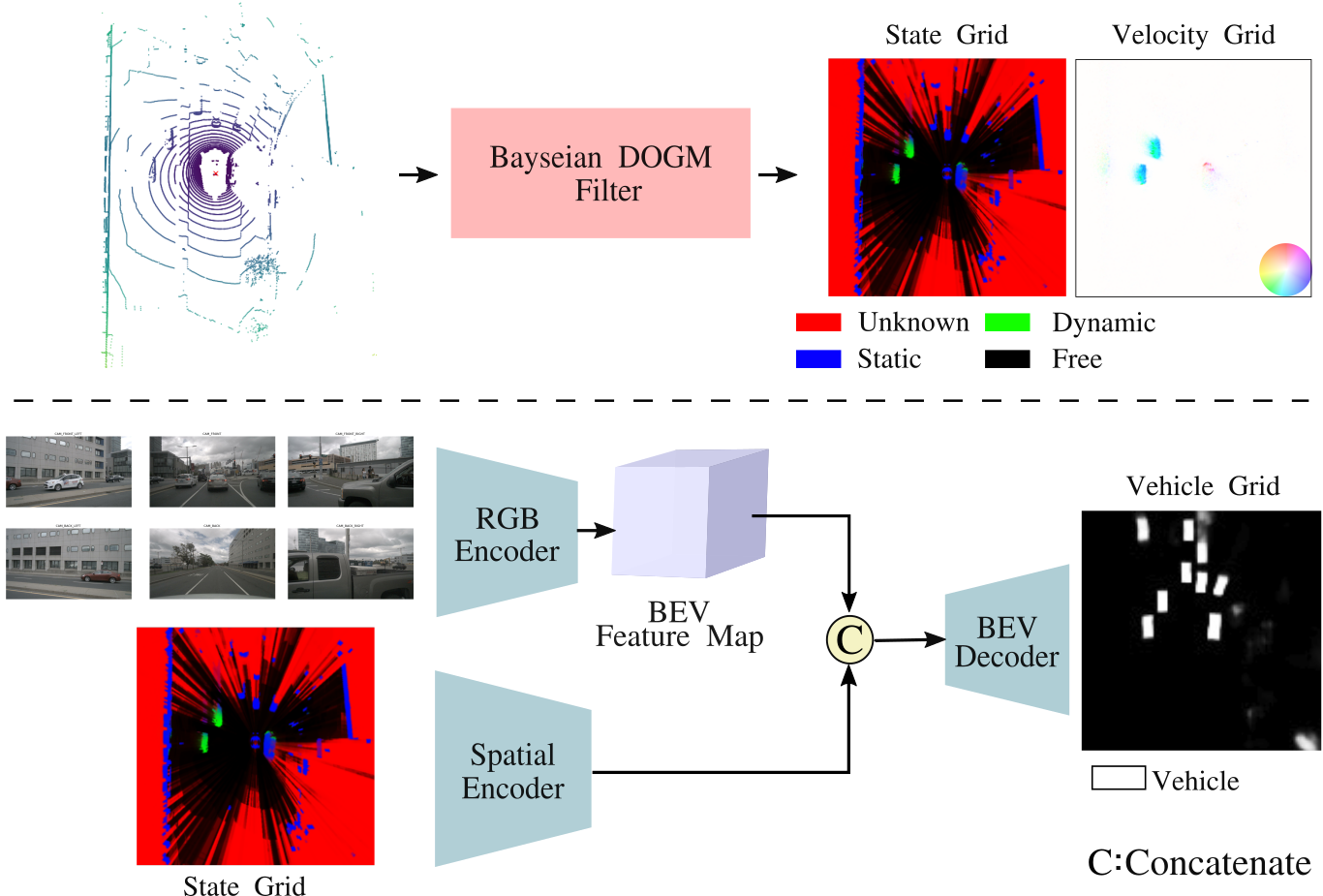}
\caption{\small Scene Representation: DOGM state grids and velocity grids $\{\mathbf{O}, \mathbf{V}\}$ are generated from LiDAR data. To incorporate semantic information, BEV features encoded from camera images are fused with occupancy state grids to predict vehicle segmentation grids $\mathbf{S}_t$.}
\label{fig:input}
\end{figure}

To complement the agent-agnostic motion cues of DOGMs, we integrate agent-specific semantic context through a vehicle segmentation grid derived from camera images. Each cell in the segmentation grid represents the probability of being occupied by a vehicle. To generate this information, we adapt the Lift, Splat, Shoot (LSS) method \cite{philion2020lift}, which extracts feature maps from multiple input camera images in BEV. Our adaptation concatenates corresponding DOGMs with the BEV feature maps and decodes the combined information into segmentation grids, Fig. \ref{fig:input}. The resulting vehicle segmentation grid $\mathbf{S}_t$ is included as the final channel in our model input.

Notably, alternative methods to include semantic information in this framework can be used. Semantic segmentation on camera images can also be fused with the Bayesian perception framework, providing a probabilistic approach to estimate vehicle segmentation grid.

The final input at time t is formed as: 
\begin{equation} \mathbf{X}_t = \text{concat}(\mathbf{O}^{\text{unk}}, \mathbf{O}^{\text{stat}}, \mathbf{O}^{\text{dyn}} , \mathbf{v}_x, \mathbf{v}_y, \mathbf{S}) \end{equation} 
By providing the model with explicit velocity and semantic priors, we reduce the search space for the subsequent prediction heads, allowing the network to focus on the high-level task of scene evolution. This structured input $\mathbf{X}_{t-N:t}$ is fed to the spatiotemporal backbone, described in the following section.

\subsection{Core Network Architecture} \label{subsec:network}
{We employ a modular pipeline of established components to treat the underlying network as a pluggable backbone. This allows us to isolate the benefits of our loss function while maintaining a path for scalability; the backbone can be easily swapped for state-of-the-art foundation models to leverage higher-capacity representations as they emerge.}
\subsubsection{Modeling Spatiotemporal Data}
Our BEV input offers a simplified scene representation, enabling the model to effectively learn the dynamics of the scene and identify possible interacting components. 
The input sequence $\mathbf{X}_{t-N:t}$ is processed through a shared spatial encoder $E_{\phi}$ to extract feature maps $\mathbf{F}_i$ for each timestep. These features are sequentially fed into a block of four ConvLSTM units, each containing 128 hidden channels  \cite{lee2021video}.  By processing the concatenated channels, the ConvLSTM learns the inter-dependencies between the occupancy states $\mathbf{O}_t$, velocity $\mathbf{V}_t$ and the vehicle segmentation $\mathbf{S}_t$ priors. This stage captures spatial and temporal dependencies, producing a compressed hidden state $\mathbf{h}_t$ and cell state $\mathbf{c}_t$ that encode the historical dynamics of the scene over the past $N$ timesteps. This hidden state $\mathbf{h}_t$ serves as the primary context for the subsequent probabilistic modeling and forecasting stages.

\subsubsection{Conditional Variational Modeling} \label{subsubsec:probab}
To account for the inherent ambiguity of future scenes, we adopt a conditional variational approach following Fiery \cite{fiery2021}, \cite{hu2020probabilistic}.
The network learns to parameterize two diagonal Gaussian distributions in a 32-dimensional latent space: a \textit{present distribution} $P_{\phi}(z \mid \mathbf{h}_t)$ conditioned on the historical hidden state that reflects the existing state of the scene, and a \textit{future distribution} $Q_{\psi}(z \mid \mathbf{h}_t, \mathbf{Y}_{1:T})$ conditioned on both history and future ground truth vehicle and flow grids. During training, the network is regularized by minimizing the Kullback-Leibler (KL) divergence $D_{\text{KL}}(Q_{\psi} \parallel P_{\phi})$ between these distributions. During inference, we draw a latent sample $z$ from $P_{\phi}$ to generate a single plausible version of the future scene.

\subsubsection{Spatiotemporal Future Predictions}
The spatial Gated Recurrent Units (sGRUs) leverage spatiotemporal features $\mathbf{h}_t$ to iteratively forecast future states, conditioned on sample $z$ drawn from learned Gaussian distributions. Each future hidden state $\mathbf{\hat{h}}_\tau$ is generated by feeding the state of the previous step $\mathbf{\hat{h}}_{\tau-1}$ into a cell comprising three convolutional GRUs and residual convolutional units \cite{ballas2015delving}. By operating on the fused spatiotemporal features, the sGRUs model the interaction between agent-specific motion and the evolving occupancy map. This ensures that the predicted features $\mathbf{\hat{h}}_{1:T}$ maintain a structural basis for temporal consistency, which we then strictly enforce through our multi-head loss functions to align the predicted evolution $\mathbf{\hat{Y}}$ with the ground truth dynamics.

\subsection{Decoder heads}
The decoder architecture consists of task-specific heads that project the predicted features $\mathbf{\hat{h}}_{1:T}$ into high-resolution grid representations. Each head employs 2D transposed convolutions to upsample the features to the original spatial resolution.

\subsubsection{Detection Head $\mathbf{Z}^{\text{det}}$ for Vehicle and Dynamic components}
This head processes the first future hidden state $\mathbf{\hat{h}}_{1}$ to predict the vehicle grid $\mathbf{D}_{\text{det}}^{\text{veh}} \in \mathbb{R}^{1 \times w \times h}$ and the agent-agnostic dynamic occupancy grid $\mathbf{D}_{\text{det}}^{\text{dyn}} \in \mathbb{R}^{1 \times w \times h}$ at the current timestep. Empirically, we found that using $\mathbf{\hat{h}}_{1}$ yields superior detection accuracy compared to the initial hidden state $\mathbf{h}_t$. This is likely because $\mathbf{\hat{h}}_{1}$ has been refined by the recurrent layers, allowing the model to better {distinguish} between static and dynamic agents. The vehicle grid $\mathbf{D}_{\text{det}}^{\text{veh}}$ captures the occupancy of both static and dynamic vehicles, allowing the network to recognize that static vehicles also influence the behavior of dynamic ones. In contrast, $\mathbf{D}_{\text{det}}^{\text{dyn}}$ identifies the dynamic components in the scene without agent semantics. While $\mathbf{X}_t$ contains much of this information, input grids are inherently noisy. This head refines these features to identify dynamic elements that might be misclassified as static due to low speed or partial occlusion. Furthermore, these grids serve as the starting point for the flow-guided predictions in Sec. \ref{subsec:flow-guided}.

\subsubsection{Prediction Head $\mathbf{Z}^{\text{pred}}$ for Vehicle and Flow Grids}
To forecast the sequence of future states, this head iteratively processes the future hidden states $\mathbf{\hat{h}}_\tau$ generated by the sGRUs for each future timestep $\tau \in \{1, \dots, T\}$. It generates a 3-channel spatial grid comprising the future vehicle grids $\mathbf{P}_{\tau}^{\text{veh}} \in \mathbb{R}^{1 \times w \times h}$ and the scene flow grids $\mathbf{P}_{\tau}^{\text{flow}} \in \mathbb{R}^{2 \times w \times h}$. By mapping the spatiotemporal features $\mathbf{\hat{h}}_\tau$ directly to this decoder, this head translates the abstract dynamics learned by the recurrent blocks into explicit motion cues used for consistency enforcement. 
The flow grid represents the evolution of occupancy using {backward flow}, a technique commonly applied in self-supervised video interpolation and optical flow tasks. At each future timestep $\tau \in \{1, \dots, T\}$, the flow grid $\mathbf{P}_{\tau}^{\text{flow}}$ encodes velocity vectors $(\mathbf{v}_x, \mathbf{v}_y)$ for each cell, indicating the displacement from the current position back to its origin at $\tau-1$, see Fig. \ref{fig:flow}. Unlike forward flow, the backward representation effectively captures diverse motion trajectories in a single-mode grid.

\subsubsection{DOGM Prediction Head $\mathbf{Z}^{\text{ogm}}$ for Occupancy States}
This head processes the future hidden states $\mathbf{\hat{h}}_\tau$ to predict the sequence of future 3-channel agent-agnostic probabilistic occupancy grids $\{\mathbf{O}^{\text{unk}}, \mathbf{O}^{\text{stat}}, \mathbf{O}^{\text{dyn}}\}$. While the input from the sGRUs primarily carries the evolution of dynamic components, the geometric context of static and unknown regions is largely preserved through skip connections from the spatial encoder $E_{\phi}$. By concatenating these high-resolution encoder features with the sGRU-derived features, the network maintains the structural integrity of the environment while also predicting the evolution of these states.

\subsection{Flow-guided Occupancy Predictions}\label{subsec:flow-guided}
In addition to direct occupancy predictions, we model scene evolution by coupling the detection head outputs $\mathbf{D}_{\text{det}}$ with predicted scene flow $\mathbf{P}_{\tau}^{\text{flow}}$. This dual-pathway approach enforces {mutual consistency} between the predicted occupancy and the scene flow through time. While the DOGM provides a probabilistic map of where objects exist, the scene flow provides the {motion cues} that dictate how those objects may move. By supervising both, the model is forced to learn a scene representation where the {spatial layout} and the {temporal evolution} are logically connected.

\begin{figure}[t]
	\centering
	    \includegraphics[trim={80 60 80 60},clip,width=1.0\columnwidth]{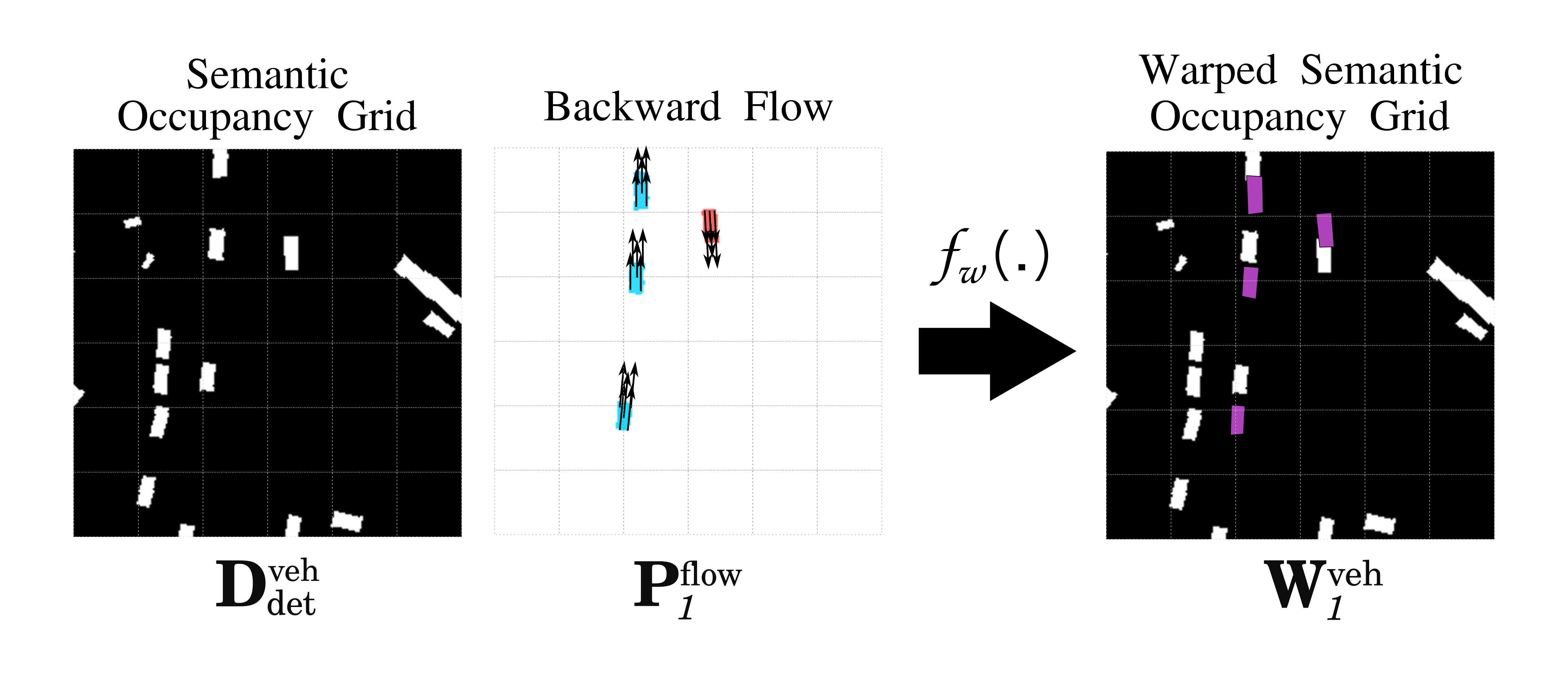}
\caption{\small Flow-guided occupancy prediction: The warping operation $f_w(\cdot)$ propagates cell occupancies (shown in purple) from the initial detection grid $\mathbf{D}_{\text{det}}^{\text{veh}}$ based on the motion vectors, that point to the origin of their respective occupancy in the backward flow grid $\mathbf{P}_{1}^{\text{flow}}$. The figure illustrates the first step ($\tau=1$) of the recursive process.}
\label{fig:flow}
\end{figure}

\textbf{Warping Occupancy with Flow: }
We leverage these grids $\mathbf{P}_{\tau}^{\text{flow}}$ to forecast the future state of occupied cells via a recursive warping operation $f_w(\cdot)$, see Fig. \ref{fig:flow}. To generate the sequence of flow-guided grids $\mathbf{W}_{1:T}$, the process is initialized with the detection grid $\mathbf{D}_{\text{det}}$. For each subsequent step, the grid is updated by displacing cell occupancies according to the motion vectors in $\mathbf{P}_{\tau}^{\text{flow}}$, such that $\mathbf{W}_{\tau} = f_w(\mathbf{W}_{\tau-1}, \mathbf{P}_{\tau}^{\text{flow}})$. Initially, $\mathbf{W}_t$ can be either the vehicle grid $\mathbf{D}_{\text{det}}^{\text{veh}}$ or the dynamic occupancy grid $\mathbf{D}_{\text{det}}^{\text{dyn}}$, resulting in warped vehicle grids $\hat{\mathbf{W}}_{1:T}^{\text{veh}}$ or warped dynamic occupancy grids $\hat{\mathbf{W}}_{1:T}^{\text{dyn}}$. 
As shown in Fig.~\ref{fig:flow}, the dynamic cells are propagated to their predicted future coordinates while the rest of the grid remains the same.

\textbf{Learning Global Scene Flow: }
$\mathbf{P}^{\text{flow}}$ in Fig.~\ref{fig:flow} represents the ground truth motion vectors specifically for dynamic components, while the majority of the grid remains empty, containing no motion vectors. In our network, we leverage the flow-guided occupancy predictions to encourage the model to infer motion vectors across \textit{all} grid cells, capturing diverse behaviors.
To this end, we employ a composite loss formulation:
\textbf{i) a flow loss }$\mathcal{L}_F$ is computed strictly over occupied cells where ground truth flow or static occupancy is present, forcing the network to learn the true velocity of dynamic agents and recognize static regions; 
\textbf{ii) a warped grid loss} $\mathcal{L}_{\mathcal{W}}$ is applied over the \textit{entire} predicted sequence $\mathbf{W}_{1:T}$ relative to the ground truth occupancy $\mathbf{Y}_{1:T}$. 
Because $\mathcal{L}_{\mathcal{W}}$ is computed over the warping operation, it allows gradients to flow back through the motion vectors of every cell. This forces the sGRUs and the prediction head to infer plausible motion patterns even for unidentified objects, ensuring a globally consistent motion field. The multi-task loss formulation is detailed in the following section.

\subsection{Losses} \label{sec:approach-losses}
The network is trained using a composite loss function designed to capture both the dynamics of the scene and the inherent uncertainty of future occupancy states. The formulation ensures that the resulting agent-specific and agent-agnostic predictions are {spatially and temporally consistent} with the predicted flow grids. 
In all following notations, a tilde $\tilde{\cdot}$ signifies ground truth for the loss calculations, and $(x, y)$ indices for grid cells are omitted for brevity.
\\
\textbf{Decoder Head Loss: }
We employ a classification loss for the detection grids using binary cross-entropy $\mathcal{H}$ over all grid cells. The detection loss is defined as:
\begin{equation}
    \mathcal{L}_{det} = \frac{1}{hw} \sum_{x=0}^{w-1} \sum_{y=0}^{h-1} \mathcal{H}(\mathbf{D}_{\text{det}}, {\mathbf{\tilde{D}}_{\text{det}}})
\end{equation}
This term is applied to both the vehicle detection $\mathbf{D}_{\text{det}}^{\text{veh}}$ and dynamic detection grids $\mathbf{D}_{\text{det}}^{\text{dyn}}$ with a weight of $\lambda_{\text{det}}$.
\\
\textbf{Prediction Head Loss: }
Similarly, the vehicle prediction sequence is supervised via classification loss:
\begin{equation}
    \mathcal{L}_{P} = \frac{1}{hwT} \sum_{\tau=1}^{T} \sum_{x=0}^{w-1} \sum_{y=0}^{h-1} \mathcal{H}(\mathbf{P}_{\tau}^{\text{veh}}, \mathbf{\tilde{P}}_{\tau}^{\text{veh}})
\end{equation}
\\
For scene flow, we use an $L_1$ regression loss. To focus the model on relevant motion, this loss is masked to include only cells containing dynamic agents or cells designated as static in the DOGM:
\begin{equation}
\begin{split}
    \mathcal{L}_F = \frac{1}{hwT} \sum_{\tau=1}^{T} \sum_{x=0}^{w-1} \sum_{y=0}^{h-1} \Big\lVert \mathbf{P}_{\tau}^{\text{flow}} - \mathbf{\tilde{P}}_{\tau}^{\text{flow}} \Big\rVert_1 \cdot \\
     \mathbb{I}(\mathbf{\tilde{P}}_{\tau}^{\text{flow}} \neq 0 \text{ or } \mathbf{\tilde{O}}_{\tau}^{\text{stat}} > 0.5)
\end{split}
\end{equation}
$\mathcal{L}_{P}$ and $\mathcal{L}_{F}$ are assigned $\lambda_{\text{veh}}$ and $\lambda_{\text{flow}}$ weights respectively.
\\
\textbf{DOGM Prediction Head Loss: }
Regression loss is employed via Mean Squared Error ($L_2$) for the occupancy probabilities of the three channels $\{\mathbf{O}^{\text{unk}}, \mathbf{O}^{\text{stat}}, \mathbf{O}^{\text{dyn}}\}$
\begin{equation}
    \mathcal{L}_{OGM} = \frac{1}{hwT} \sum_{\tau=1}^{T} \sum_{x=0}^{w-1} \sum_{y=0}^{h-1} \Big\lVert \mathbf{O}_{\tau}^{\text{}} - \mathbf{\tilde{O}}_{\tau}^{\text{}} \Big\rVert_2
\end{equation}
Channel losses are weighted by $\lambda_{\text{unk}}$, $\lambda_{\text{stat}}$ and $\lambda_{\text{dyn}}$.
\\
\textbf{Flow-Guided Prediction Loss: }
We incorporate the flow-traced loss \cite{mahjourian2022occupancy}. Warped vehicle grids $\mathbf{W}_{\tau}^{\text{veh}}$ and warped dynamic occupancy grids $\mathbf{W}_{\tau}^{\text{dyn}}$ are weighted with vehicle prediction grids $\mathbf{P}^{\text{veh}}$ and DOGM dynamic grids $\mathbf{O}^{\text{dyn}}$ respectively. This helps to jointly learn occupancy and flow for future timesteps based on detected vehicles and agent-agnostic dynamic entities. Loss function is defined as follows:
\begin{equation*}
    \mathcal{L}_{\mathcal{W}}^{\text{veh}} = \frac{1}{hwT} \sum_{\tau=1}^{T} \sum_{x=0}^{w-1} \sum_{y=0}^{h-1} \mathcal{H}(\mathbf{W}_{\tau}^{\text{veh}} \cdot \mathbf{P}_{\tau}^{\text{veh}}, \mathbf{\tilde{P}}_{\tau}^{\text{veh}})
\end{equation*}
\begin{equation}
    \mathcal{L}_{\mathcal{W}}^{\text{dyn}} = \frac{1}{hwT} \sum_{\tau=1}^{T} \sum_{x=0}^{w-1} \sum_{y=0}^{h-1} \mathcal{H}(\mathbf{W}_{\tau}^{\text{dyn}} \cdot \mathbf{O}_{\tau}^{\text{dyn}}, \mathbf{\tilde{O}}_{\tau}^{\text{dyn}})
\end{equation}
$\mathcal{L}_{\mathcal{W}}^{\text{veh}}$ and $\mathcal{L}_{\mathcal{W}}^{\text{dyn}}$ are assigned $\lambda_{\mathcal{W}_{\text{veh}}}$ and $\lambda_{\mathcal{W}_{\text{dyn}}}$ weights respectively.

Finally, the Kullback-Leibler divergence $\mathcal{L}_{KL}$ is introduced to regularize the latent space. The total loss is the weighted sum of all components:
\begin{equation}
\begin{split}
\mathcal{L} = & \lambda_{\text{det}} \mathcal{L}_{\text{det}} + \lambda_{\text{veh}}\mathcal{L}_{P} + \lambda_{\text{flow}}\mathcal{L}_{F} + \lambda_{kl}\mathcal{L}_{KL} + \\
& + \sum_{c \in \{\text{unk, stat, dyn}\}} \lambda_c \mathcal{L}_{OGM}^c + \lambda_{\mathcal{W}_{\text{veh}}}\mathcal{L}_{\mathcal{W}}^{\text{veh}} + \lambda_{\mathcal{W}_{\text{dyn}}}\mathcal{L}_{\mathcal{W}}^{\text{dyn}}
\end{split}
\end{equation}
The balancing of these loss components via task-specific weights is critical. With scene flow grids as the core, prioritizing $\lambda_{\text{flow}}$ improves the spatial accuracy of vehicle and dynamic grid predictions, enhancing overall scene consistency.
\section{EXPERIMENTS} \label{sec:experiments}
We evaluate our proposed model on two publicly available real-world datasets: nuScenes \cite{nuscenes2019} and the Woven Planet Perception dataset \cite{WovenPlanet2019}. Both datasets provide raw onboard LiDAR and camera sensor data, along with refined ego-vehicle pose, allowing generation of DOGMs and vehicle segmentation grids.
The nuScenes dataset comprises 700 training and 150 validation scenes, each 20s long, yielding a total of 280 minutes of annotated urban driving data in Singapore and USA. The 32-layer LiDAR sensor data is provided at a frequency of 10Hz, with annotated keyframes at 2Hz.
The Woven Planet (WP) dataset provides 64-layer LiDAR point cloud at 5 Hz, all with annotated keyframes. This dataset, collected in USA, comprises of 180 scenes, each 25s long, totaling 75 minutes of annotated driving data. Both datasets provide 360-degree coverage around the vehicle with 6 camera images. While both focus on urban environments, the WP dataset features denser traffic with higher vehicle count, whereas nuScenes captures a broader range of driving behaviors.

\subsection{Dataset Preparation}
LiDAR-based DOGMs are generated, comprising state and velocity grids $\{\mathbf{O}, \mathbf{V}\}$ with a resolution of $0.1\text{m}$. For vehicle grids $\mathbf{S}$ preparation, all 6 camera images are used in conjunction with the corresponding DOGM state grids. Ground truth annotations for the future vehicle grids $\mathbf{P}^{\text{veh}}$ are derived from the available vehicle bounding box annotations. Note that the ego-vehicle is not included in the vehicle grids. The ground truth flow $\mathbf{P}^{\text{flow}}$ of the agents is estimated by calculating the displacement of agent centroids on the grid. Additionally, the future DOGM channels \{${\mathbf{O}^{\text{stat}}, \mathbf{O}^{\text{dyn}}}$\} are adjusted using the ground truth annotations to ensure that static components are not represented as dynamic occupancies, and vice versa. 

All grids are generated from an ego-centric perspective, with the ego-vehicle located at the center and oriented upwards. Future predictions for each sequence are made relative to the ego-vehicle frame of reference at the current timestep. To facilitate allo-centric predictions in a fixed frame, the past and future grids are transformed using the available odometry in the dataset and are then cropped to cover an area of $60$x$60\text{m}^2$.
In total, the nuScenes dataset contains 23K training and 5K validation sequences wheres as WP dataset contains 7K training and 2K validation sequences.

\subsection{Training setup}
The input sequence $\mathbf{X}_{t-2:t}$ consists of 3 frames, spanning over $1.0$s. The network is trained to predict the present detection head grids $\mathbf{Z}^{\text{det}}$ and $5$ future frames with steps of $0.5$s for both prediction $\mathbf{Z}_{1:5}^{\text{pred}}$ and DOGM head $\mathbf{Z}_{1:5}^{\text{ogm}}$. During training, the grids are resized to $240$x$240$ cells, resulting in a spatial resolution of $0.25\text{m}$ per cell. This choice follows common benchmarks ranging from $0.2$-$0.5\text{m}$ for larger domains and prioritizes comprehensive situational awareness over the entire scene. Furthermore, the inclusion of scene flow allows the model to capture sub-pixel motion, mitigating the limitations of grid discretization.

\textbf{Implementation details:}
The network is implemented in PyTorch and trained on 8 Nvidia Tesla V100 GPUs. We use the Adam optimizer with a learning rate of 3x${10}^{-4}$ and a weight decay of $3$x${10}^{-7}$. The model is trained for $20$ epochs with a batch size of 18, taking approximately $5$ hours to complete for nuScenes and 2 hours for WP dataset. 

\textbf{Selecting loss coefficients:}
The loss weights are empirically balanced to prioritize the spatiotemporal accuracy of the entire prediction sequence. Rather than optimizing each decoder head in isolation, the coefficients are tuned to ensure the network extracts robust latent features that ensure more reliable and consistent occupancy forecasting across the $2.5$s horizon. The loss coefficients are set to 
$\lambda_{\text{det}}$=$0.25$, $\lambda_{\text{veh}}$=$1.0$, 
$\lambda_{\text{unk}}$=$\lambda_{\text{stat}}$=$1.0$, $\lambda_{\text{dyn}}$=$6.0$, $\lambda_{\mathcal{W}_{\text{veh}}}$=$0.1$, $\lambda_{\mathcal{W}_{\text{dyn}}}$=$0.01$ and $\lambda_{kl}$=$0.005$.
The flow grid coefficient $\lambda_{\text{flow}}$ differs for the two datasets. Due to numerous occlusions in the WP dataset, a higher weight of $50.0$ improves performance. In contrast, the best performances on the nuScenes are achieved with a coefficient of $10.0$.

\textbf{Runtime performance: }
The proposed framework comprises approximately $9.4$M parameters and achieves a prediction network inference latency of $35$ms on one V100 GPU. The input preparation combines a CUDA-optimized DOGM module ($5$ms) and an LSS-based vehicle segmentation module ($40$ms), resulting in an end-to-end latency of $80$ms and supporting real-time operation at $10$Hz.
Unlike object-based tracking methods that suffer from increased latency in dense traffic, this grid-based representation ensures stable, scalable performance as agent density increases. Notably, the multi-decoder architecture remains efficient at $5$ms, while flow-guided occupancy warping adds a marginal $5$ms overhead.

\subsection{Evaluation baselines}
Our method is compared against two baseline video prediction networks that have been widely adopted in recent OGM forecasting literature \cite{toyungyernsub2024predicting, Toyungyernsub2022, mann2022predicting,itkina2019dynamic}: \textit{PredNet} \cite{lotter2016deep}, a Predictive Coding Network that comprises of vertically-stacked ConvLSTMs and \textit{PredRNN} \cite{wang2022predrnn}, which employs memory-decoupled spatio-temporal LSTM blocks, with zigzag memory flow and a curriculum learning strategy.
Unlike the multi-decoder heads of our model, the baseline networks predict the same set of channels in the output as in the input. Both networks are trained for a 4-channel configuration comprising of vehicle grid and DOGM grids. \textit{PredRNN} is additionally trained for a 6-channel configuration that also includes the velocity input and future flow grids. The input sequence and output ground truth frames are the same for our proposed and baseline models.

\subsection{Evaluation Metrics}
We assess the performance of the grids using grid-based metrics. The following metrics are used:\\
1. \textit{Intersection Over Union (IoU)} measures the ratio of correctly predicted occupied cells to the total number of predicted and ground truth occupied cells. IoU typically considers binary classification by thresholding predicted values. In contrast, soft-IoU accounts for the probability value of predictions, reflecting the model confidence for each cell.\\
2. \textit{Recall} measures the ratio of correctly predicted occupied cells to the total ground truth occupied cells. Similar to IoU, standard recall uses binary classification, while soft-recall incorporates the probability values of the predicted cells.\\
3. \textit{Recall-Dynamic} computes recall specifically for dynamic agents by considering only the segmentation of dynamic agents in the ground truth grid.\\
4. \textit{End-Point Error (EPE)} measures the average Euclidean distance between the predicted and ground truth flow vectors, considering only the cells that are occupied by dynamic agents.\\
5. \textit{Mean Square Error (MSE)} measures the average squared difference between predicted and ground truth values for each cell across all channels of the DOGM grid.

Together, these metrics offer a multi-dimensional evaluation approach motivated by the need for {physical realism} and {structural integrity}. While IoU is a standard measure of spatial overlap, it penalizes diverse, feasible occupancy predictions that do not perfectly align with a single ground truth realization. In such scenarios, {Recall} provides a valuable measure of the model's ability to \textit{cover} the ground-truth occupancy among various predicted possibilities, without being punished for its diverse explorations. Meanwhile, {EPE} serves as a critical check for {physical realism}, ensuring that the generated motion is kinematically valid and follows realistic motion constraints. Finally, {MSE} provides an assessment of {structural integrity} by quantifying how well the model preserves the probabilistic distribution across all DOGM channels.

\textit{Evaluation Challenges:} The evaluation of BEV grid predictions without instance information presents notable limitations, as it does not allow for the assessment of individual agents. Furthermore, holistic scene prediction is often penalized by new occupancy or agents entering the field of view, which are inherently challenging to anticipate. Additionally, collective grid metrics can be misleading in scenes dominated by static objects; for example, a high global IoU may mask poor performance on dynamic vehicles if they represent a small fraction of the total occupancy.

\subsection{Evaluation methodology}
We report the performance of vehicle grids $\mathbf{P}_{t}^{\text{veh}}$ using soft-IoU, soft-recall and soft-recall-dynamic. The flow grids $\mathbf{P}_{t}^{\text{flow}}$ and DOGM grids $\mathbf{Z}_{t}^{\text{ogm}}$ are evaluated with EPE and MSE respectively.
To evaluate the flow-warped grids, we use IoU, recall and recall-dynamic. Warped grids typically have high probability values at each timestep since the occupancy is propagated from detection grids with the aid of flow grids. Thus, for fair evaluation, binary classification methods are selected.
Additionally, we evaluate i) the flow-warped grid weighted by the vehicle grid $\mathbf{W}_{t}^{\text{veh}}  \mathbf{P}_{t}^{\text{veh}}$, ii) the flow-warped grid itself $\mathbf{W}_{t}^{\text{veh}}$ at every future timesteps.
$\mathbf{W}_{t}^{\text{veh}}$ may contain occupancy predictions from previous timesteps and diverse predictions for future, while $\mathbf{W}_{t}^{\text{veh}}  \mathbf{P}_{t}^{\text{veh}}$ evaluates the joint performance of occupancy and flow predictions.\\
To assess agnostic prediction capabilities, we consider unrecognized agents in the scene,i.e., pedestrians and cyclists. We evaluate dynamic pedestrian and cyclist predictions in $\mathbf{O}_{t}^{\text{dyn}}$ and $\mathbf{W}_{t}^{\text{dyn}}$ grids with the recall metric.\\
For all inferences, we use the mean value of the \textit{present Gaussian distribution}, discussed in sec. \ref{subsec:network}
\section{EVALUATION} \label{sec:evaluation}
\begin{figure*}[ht]
\centering
    \includegraphics[trim={0 0 0 0},clip,scale=0.5,width=2.0\columnwidth]{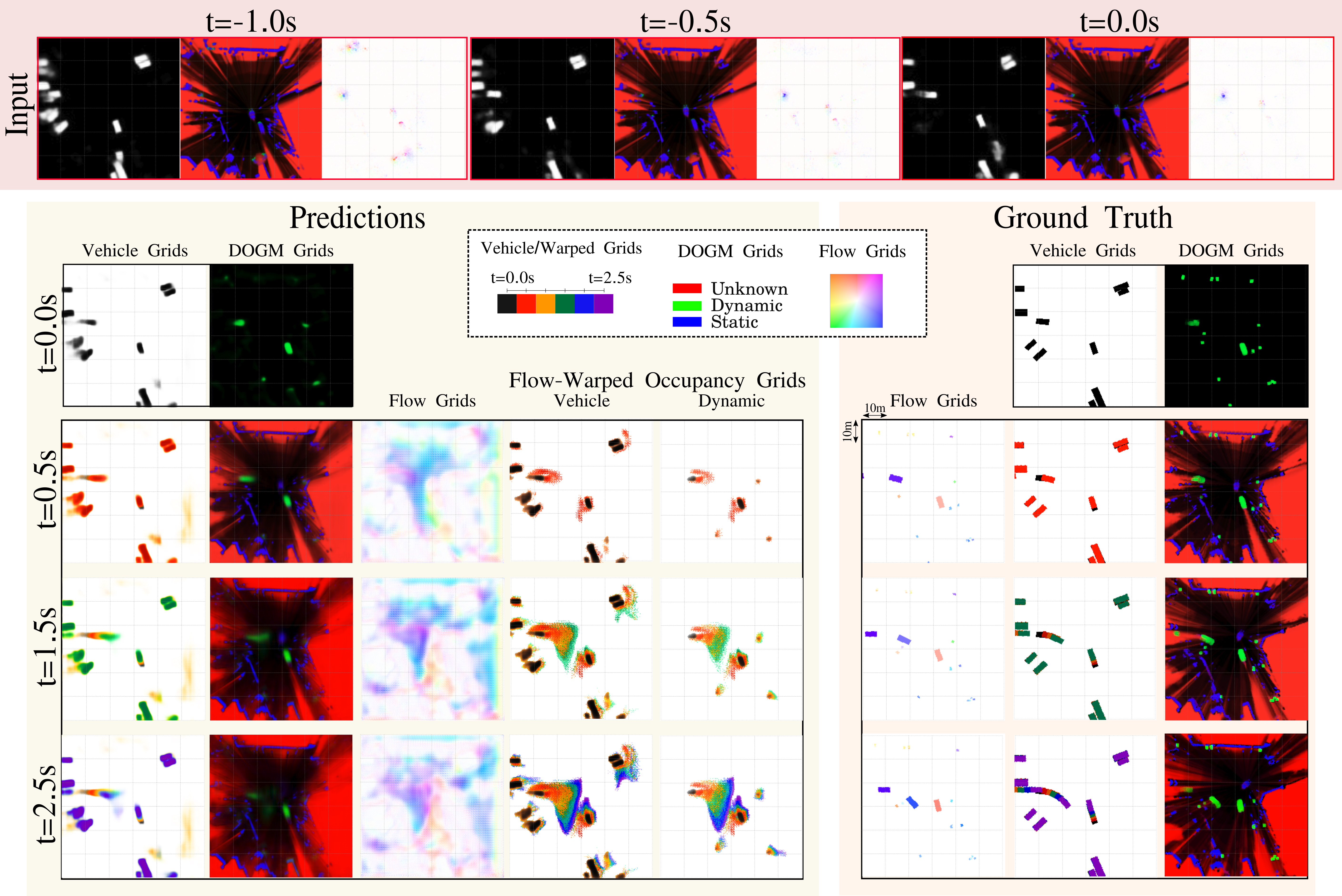}
\caption{\small Prediction example of a driving scene from the Nuscenes Datatset. The scene comprises two dynamic vehicles, multiple pedestrians and the ego-vehicle in the grid center. Input sequence comprises of the vehicle grids, DOGM state and velocity grids. Prediction results are displayed in the left five columns while right three show the corresponding ground truth grids. Detection head $\mathbf{Z}^{\text{det}}$ grids are displayed for the current timestep (t=0.0s) row while the remaining rows show prediction head $\mathbf{Z}^{\text{pred}}$ and DOGM prediction head $\mathbf{Z}^{\text{ogm}}$ grids.}
\label{fig:qualitative}
\end{figure*}
\begin{figure}[t]
	\centering
	    \includegraphics[trim={0 0 0 0},clip,width=0.9\columnwidth]{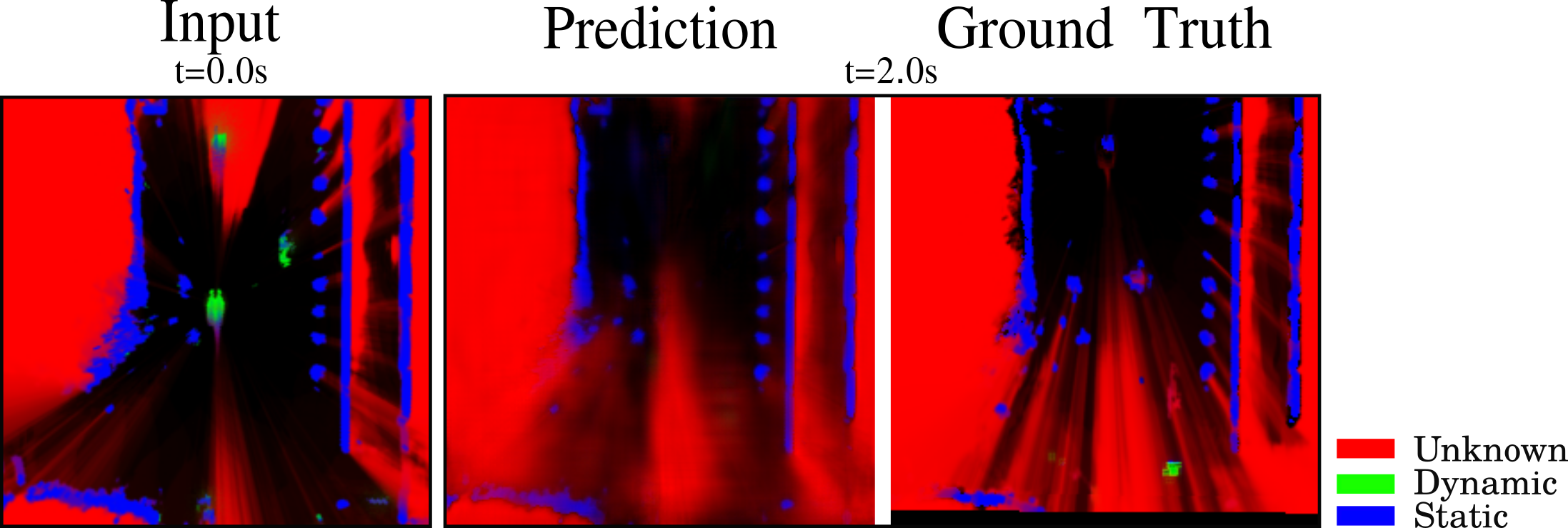}
\caption{\small Scenario illustrating prediction of the future unknown state grid for a fast ego-vehicle on the Nuscenes Datatset.}
\label{fig:unknown}
\end{figure}
\subsection{Qualitative Evaluation}
Figure \ref{fig:qualitative} shows qualitative results at a T-intersection in nuScenes dataset using our model. The vehicle grids and the velocity grids capture the motion of two dynamic vehicles in the scene input while the occupancy grids capture the scene structure as well as various pedestrians, without any semantic information associated with them. The future vehicle and DOGM grids show how the predictions of the dynamic vehicle on the left get blurry and uncertain with time as the vehicle can continue straight or take a turn. With the help of detection and flow grids, flow-warped grids show diverse possible propagation of vehicle and agent-agnostic occupancies (i.e., pedestrians in this case). For more examples and animated results, the reader can refer to the weblink\footnote{github.com/rabbia-asghar/dogm-veh-flow}. We summarize below contributions of various components in our proposed framework.\\
\textbf{Significance of Vehicle prediction}
Prediction of agent-specific grid significantly enhances the model's ability to predict both static and dynamic vehicles. The results demonstrate that the network learns the scene structure from DOGM and dynamic vehicles behavior in the scene, guiding the model in the absence of explicit road information.\\
\textbf{Significance of DOGM prediction} The DOGM predictions capture the evolution of all three occupancy states, $\{\mathbf{O}^{\text{unk}}, \mathbf{O}^{\text{stat}}, \mathbf{O}^{\text{dyn}}\}$.
Predicting static and unknown regions is challenging as their evolution largely depends on the ego-vehicle motion and the unfolding environment in the subsequent timesteps. 
Fig. \ref{fig:unknown} illustrates the evolution of the {unknown} state $\mathbf{O}^{\text{unk}}$ for a rapidly shifting ego-perspective. As the ego-vehicle advances, the model effectively predicts areas that become partially or completely unobservable due to sensor range or environmental occlusions. 
Despite becoming blurrier over time due to the added uncertainty at each timestep, these predictions are still crucial for risk estimation and safe navigation as they anticipate new occupancies and unobservable areas in the scene.\\
\textbf{Significance of Flow prediction} With DOGM input and predictions, the flow predictions $\mathbf{P}^{\text{flow}}$ ensure behavior predictions that respect the  scene structure and  maintain collision avoidance capabilities, even without explicit semantic information. Furthermore, backward flow prediction enables the network to learn diverse behaviors.\\
\textbf{Agent-agnostic dynamic predictions} The prediction of $\mathbf{D}_{\text{det}}^{\text{dyn}}$ helps the network identify the dynamic components at current timestep. With vehicles considered in the semantic grid, pedestrians offer the possibility of evaluating agent-agnostic dynamic predictions in our real-world dataset. Due to the small size and slow motion of pedestrian occupancy in the grid, they tend to soon vanish in the DOGM dynamic state grid while flow-warped dynamic occupancy grid ${\mathbf{W}}_{1:T}^{\text{dyn}}$ provide their possible evolution in the scene.

\textbf{Limitations and Failure Analysis} 
The ground truth DOGM grids Fig. \ref{fig:qualitative} reveals that our model misses some dynamic pedestrian detections since their limited cell count in the occupancy grid makes consistent detection challenging. We also note that the flow grids also assign motion vectors to static vehicles, as the network is encouraged to learn potential motion for the entire grid. Finally, the predicted flow lacks explicit kinematic constraints and without a vehicle-specific motion model, vectors may occasionally represent physically unfeasible transitions \cite{girase2021physically}. Addressing these challenges remains an objective for future work.

\subsection{Quantitative Evaluation}
We evaluate the performance of vehicle prediction, scene flow and DOGM prediction output in Table \ref{table:results-a} for both nuScenes and WP dataset. All metrics scores are averaged over 2.5s prediction horizon. Among both the datasets, we note that while our framework does not achieve the best soft-IoU, high soft-recall indicates that our model is better at predicting vehicle motion. The trade-off between IoU and recall highlights the method's tendency to overestimate occupancy, potentially coming from diverse predictions. Our model consistently excels in dynamic vehicle prediction, both in soft-recall and end-point error. For DOGM predictions, PredNet and PredRNN excel in unknown and static channel predictions showing that the video prediction networks perform better in retaining static and unknown scene information and predicting their evolution. In terms of dynamic channel, our model exhibits the lowest MSE in nuScenes dataset, that contains more diverse behaviours while for more linear behaviour prone WP dataset, the PredRNN with 4 channels performs better.\\
\begin{table*}[htbp]
    \caption{Performance comparison between networks on nuScenes and Woven Planet (WP) datasets.}    \centering
    \begin{tabular}{c|c c|c c|c c c}
        \hline
        \multicolumn{8}{c}{\textbf{nuScenes Dataset}} \\
        \hline
        & \multicolumn{2}{c|}{\textbf{All Vehicles}} & \multicolumn{2}{c|}{\textbf{Dynamic Vehicles}} & \multicolumn{3}{c}{\textbf{DOGM MSE (x${10}^{-2}$)}} \\
        \textbf{Network} & \textbf{soft-IoU} $(\uparrow)$ & \textbf{soft-recall}  $(\uparrow)$ & \textbf{soft-recall}  $(\uparrow)$ & \textbf{EPE} $(\downarrow)$ & \textbf{unknown} $(\downarrow)$& \textbf{dynamic} $(\downarrow)$ & \textbf{static} $(\downarrow)$\\
        \hline
        PredNet (4 ch) & \textbf{0.440}  & 0.514 & 0.177  & -    & 3.13  & 0.297 & 0.895 \\
        PredRNN (4 ch) & 0.397 & 0.599 & 0.444  & -    & \textbf{2.07}  & 0.235 & \textbf{0.779} \\
        PredRNN (6 ch) & 0.415 & 0.584 & 0.423  & 0.077 & 2.10   & 0.236 & 0.782 \\
        Ours           & 0.365 & \textbf{0.670}  & \textbf{0.541}  & \textbf{0.040} & 2.93  & \textbf{0.232} & 0.831 \\
        \hline
        \multicolumn{8}{c}{\textbf{WP Dataset}} \\
        \hline
        & \multicolumn{2}{c|}{\textbf{All Vehicles}} & \multicolumn{2}{c|}{\textbf{Dynamic Vehicles}} & \multicolumn{3}{c}{\textbf{DOGM MSE (x${10}^{-2}$)}} \\
        \textbf{Network} & \textbf{soft-IoU} $(\uparrow)$ & \textbf{soft-recall}  $(\uparrow)$ & \textbf{soft-recall}  $(\uparrow)$ & \textbf{EPE} $(\downarrow)$ & \textbf{unknown} $(\downarrow)$& \textbf{dynamic} $(\downarrow)$ & \textbf{static} $(\downarrow)$\\
        \hline
        PredNet (4 ch) & \textbf{0.493} & 0.564 & 0.382  & -    & \textbf{0.79}  & 0.884 & 1.11 \\
        PredRNN (4 ch) & 0.446 & 0.594 & 0.503  & -   & 0.822 & \textbf{0.512} & 0.98 \\
        PredRNN (6 ch) & 0.466 & 0.639 & 0.557  & 0.085 & 0.775 & 0.523 & \textbf{0.959} \\
        Ours           & 0.408 & \textbf{0.655} & \textbf{0.578}  & \textbf{0.030}  & 1.74  & 0.562 & 1.12 \\
        \hline
        \multicolumn{8}{c}{Legend: ch = channels, IoU = Intersection Over Union, EPE = End-Point-Error, MSE = Mean Square Error}
    \end{tabular}
\label{table:results-a}
\end{table*}
\begin{table}[h]
\caption{Vehicle Prediction Evaluation As Binary Classification.}
\centering
\begin{tabular}{c|c|c|l|c|l|l|}
\cline{1-6}
\hline
\multicolumn{6}{c}{\textbf{nuScenes Dataset}}\\\hline
\multicolumn{1}{c|}{\textbf{Network}} & \multicolumn{1}{c}{\textbf{IoU} $(\uparrow)$} & \multicolumn{2}{c}{\textbf{Recall} $(\uparrow)$}  & \multicolumn{2}{c}{\textbf{Recall-Dynamic} $(\uparrow)$} \\\hline
\multicolumn{1}{c|}{PredNet} & \multicolumn{1}{c}{{0.444}} & \multicolumn{2}{c}{0.518} & \multicolumn{2}{c}{0.181} \\ 
\multicolumn{1}{c|}{PredRNN (4 ch)} & \multicolumn{1}{c}{\textbf{0.496}} & \multicolumn{2}{c}{0.617} & \multicolumn{2}{c}{0.443}\\ 
\multicolumn{1}{c|}{PredRNN (6 ch)} & \multicolumn{1}{c}{{0.492}} & \multicolumn{2}{c}{0.604} & \multicolumn{2}{c}{0.426}\\ 
\hline
\multicolumn{1}{c|}{Ours $\mathbf{P}^{\text{veh}}$} & \multicolumn{1}{c}{0.467} & \multicolumn{2}{c}{{0.699}}  & \multicolumn{2}{c}{{0.563}}\\ 
\multicolumn{1}{c|}{Ours $\mathbf{W}^{\text{veh}}  \mathbf{P}^{\text{veh}}$} & \multicolumn{1}{c}{0.480} & \multicolumn{2}{c}{{0.662}} & \multicolumn{2}{c}{{0.499}} \\ 
\multicolumn{1}{c|}{Ours $\mathbf{W}^{\text{veh}}$} & \multicolumn{1}{c}{0.266} & \multicolumn{2}{c}{\textbf{0.734}} & \multicolumn{2}{c}{\textbf{0.657}} \\ 
\hline
\multicolumn{6}{c}{\textbf{WP Dataset}}\\\hline
\multicolumn{1}{c|}{\textbf{Network}} & \multicolumn{1}{c}{\textbf{IoU} $(\uparrow)$} & \multicolumn{2}{c}{\textbf{Recall} $(\uparrow)$}  & \multicolumn{2}{c}{\textbf{Recall-Dynamic} $(\uparrow)$} \\\hline
\multicolumn{1}{c|}{PredNet} & \multicolumn{1}{c}{{0.496}} & \multicolumn{2}{c}{0.569} & \multicolumn{2}{c}{0.385} \\ 
\multicolumn{1}{c|}{PredRNN (4 ch)} & \multicolumn{1}{c}{{0.534}} & \multicolumn{2}{c}{0.618} & \multicolumn{2}{c}{0.515}\\ 
\multicolumn{1}{c|}{PredRNN (6 ch)} & \multicolumn{1}{c}{\textbf{0.544}} & \multicolumn{2}{c}{0.659} & \multicolumn{2}{c}{{0.571}}\\ 
\hline
\multicolumn{1}{c|}{Ours $\mathbf{P}^{\text{veh}}$} & \multicolumn{1}{c}{0.505} & \multicolumn{2}{c}{{0.689}}  & \multicolumn{2}{c}{{0.596}}\\ 
\multicolumn{1}{c|}{Ours $\mathbf{W}^{\text{veh}}  \mathbf{P}^{\text{veh}}$} & \multicolumn{1}{c}{0.506} & \multicolumn{2}{c}{{0.643}} & \multicolumn{2}{c}{{0.547}} \\ 
\multicolumn{1}{c|}{Ours $\mathbf{W}^{\text{veh}}$} & \multicolumn{1}{c}{0.282} & \multicolumn{2}{c}{\textbf{0.748}} & \multicolumn{2}{c}{\textbf{0.704}} \\ 
\hline
\multicolumn{6}{c}{Legend: ch = channels, IoU = Intersection Over Union}
\end{tabular}
\label{table:results-warped}
\end{table}
\begin{figure}[t]
	\centering
	    \includegraphics[trim={0 0 0 0},clip,width=0.8\columnwidth]{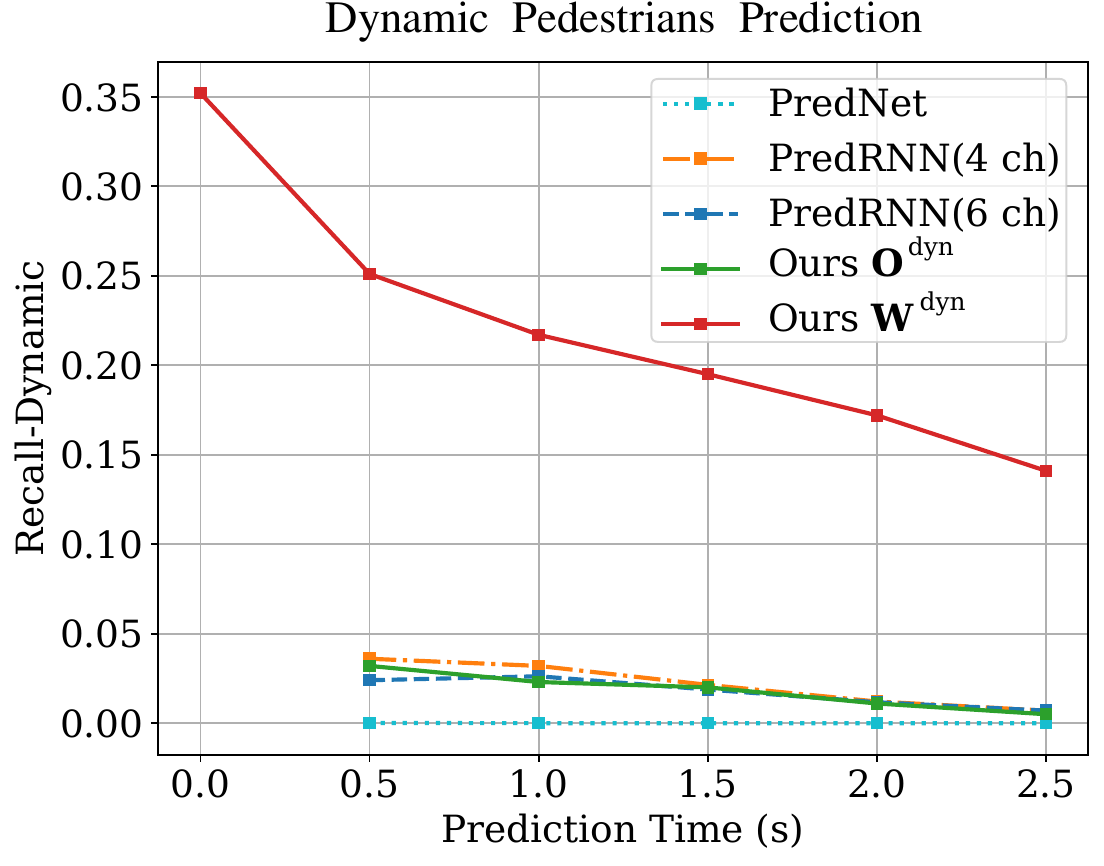}
\caption{\small Plot for dynamic pedestrians' to highlight the agent-agnostic prediction capabilities on nuScenes dataset. Recall for dynamic pedestrians is computed  for DOGM $\mathbf{O}^{\text{dyn}}$  across different prediction networks along with warped dynamic grids $\mathbf{W}^{\text{dyn}}$ in our network.}
\label{fig:plot}
\end{figure}
\begin{table*}[h]
\centering
\caption{Ablation study on different prediction configurations on nuScenes Dataset.}    \centering
\begin{tabular}
{>{\centering\arraybackslash}p{0.2cm}>{\centering\arraybackslash}p{.4cm}|>{\centering\arraybackslash}p{.2cm}>{\centering\arraybackslash}p{.2cm}>{\centering\arraybackslash}p{.4cm}|>{\centering\arraybackslash}p{.4cm}>{\centering\arraybackslash}p{.2cm}|
>{\centering\arraybackslash}p{.8cm}|
>{\centering\arraybackslash}p{.5cm}
>{\centering\arraybackslash}p{.5cm}
>{\centering\arraybackslash}p{.5cm}|
>{\centering\arraybackslash}p{.6cm}
>{\centering\arraybackslash}p{.6cm}
>{\centering\arraybackslash}p{.8cm}|>{\centering\arraybackslash}p{.6cm}>{\centering\arraybackslash}p{.6cm}>{\centering\arraybackslash}p{.6cm}}
\hline
\multicolumn{5}{c|}{\textbf{Decoder Output}} & \multicolumn{2}{c|}{\textbf{Warped}} & & \multicolumn{3}{c|}{} & \multicolumn{3}{c|}{} & \multicolumn{3}{c}{\textbf{Warped}}  \\
\multicolumn{5}{c|}{\textbf{Configuration}} & \multicolumn{2}{c|}{\textbf{Grids Loss}} & \textbf{Flow}$(\downarrow$) & \multicolumn{3}{c|}{\textbf{MSE (x${10}^{-2}$)} $(\downarrow$)} & \multicolumn{3}{c|}{\textbf{Vehicles}$(\uparrow$)} & \multicolumn{3}{c}{\textbf{Recall Dynamic}$(\uparrow)$}  \\
\hline
{$\mathbf{D}^{\text{veh}}$} & $\mathbf{D}^{\text{dyn}}$ & $\mathbf{P}^{\text{veh}}$ & $\mathbf{P}^{\text{flow}}$ & $\mathbf{Z}^{\text{ogm}}$ & $\mathbf{W}^{\text{veh}}$ & {$\mathbf{W}^{\text{dyn}}$} & \textbf{EPE} & $\mathbf{O}^{\text{unk}}$ & $\mathbf{O}^{\text{dyn}}$ & $\mathbf{O}^{\text{stat}}$ & \textbf{IoU} & \textbf{Recall} & \textbf{R-dyn} & \textbf{Veh} & \textbf{Ped} & \textbf{Cyc} \\
\hline
&  & \checkmark & \checkmark &  &  &  & \textbf{0.039} & - & - & - & 0.466 & \textbf{0.714} & \textbf{0.601} & - & - & -\\
&  & \checkmark &  & \checkmark &  &  & - & 2.95 & \underline{0.241} & 0.838 & 0.468 & 0.698 & \underline{0.569} & - & - & -\\
&  & \checkmark & \checkmark & \checkmark &  &  & 0.042 & \underline{2.79} & 0.324 & \underline{0.826} & \underline{0.477} & 0.698 & 0.562 & - & - & -\\
\checkmark &  & \checkmark & \checkmark &  & \checkmark &  & 0.041 & - & - & - & 0.466 & 0.677 & 0.516 & \textbf{0.658} & - & -  \\
\checkmark & \checkmark & \checkmark & \checkmark & \checkmark &  &  & \underline{0.040} & \textbf{2.69} & 0.324 & \textbf{0.823} & \textbf{0.480} & 0.696 & 0.557 & 0.621 & \underline{0.016} & \underline{0.105}\\
\hline
\checkmark & \checkmark & \checkmark & \checkmark & \checkmark & \checkmark & \checkmark & \underline{0.040} & 2.93 & \textbf{0.232} & 0.831 & 0.467 & \underline{0.699} & 0.563 & \underline{0.657} & \textbf{0.18} & \textbf{0.168}\\
\hline
\multicolumn{16}{c}{Legend: EPE = End-Point-Error, MSE = Mean Square Error,  IoU = Intersection Over Union, R-dyn = Recall-Dynamic}\\
\multicolumn{10}{c}{ Veh = Vehicles, Ped = Pedestrians, Cyc = Cycles }
\end{tabular}
\label{tab:ablation-study}
\end{table*}
\\
\textbf{Flow-guided agent predictions} Table \ref{table:results-warped} presents the vehicle prediction results, averaged across the 2.5s prediction horizon. On both datasets, the warped vehicle grids $\mathbf{W}^{\text{veh}}$ have significantly lower IoU since they offer very diverse flow of predictions and also have the tendency to retain occupancy from previous timesteps. However, when weighted by the prediction grid $\mathbf{W}^{\text{veh}}  \mathbf{P}^{\text{veh}}$, the IoU improves. While the true positive occupancies reduce in  $\mathbf{W}^{\text{veh}}  \mathbf{P}^{\text{veh}}$, the false positives also get eliminated. Compared to vehicle grids $\mathbf{P}^{\text{veh}}$, recall and recall-dynamic of warped grids $\mathbf{W}^{\text{veh}}$ significantly improve, by 5\% and 31\% percent in nuScenes and 9\% and 18\% in WP.\\
\textbf{Agent-agnostic prediction capabilities} Fig. \ref{fig:plot} illustrates the network capabilities to predict dynamic pedestrians in the nuScenes dataset. Without semantic information,
pedestrian motion is easier to miss and their occupancy recall rapidly drops to zero. The plotline \textit{Ours} $\mathbf{W}^{\text{dyn}}$ begins with the recall of dynamic pedestrians in $\mathbf{D}^{\text{dyn}}$ at t=0 and then continues with flow-guided occupancy prediction over next timesteps. It is clear that warped dynamic occupancy grids significantly outperform $\mathbf{O}^{\text{dyn}}$ in capturing possible agent-agnostic motion. \\

\textbf{Ablation Study} 
Table \ref{tab:ablation-study} evaluates different decoder configurations. Our complete model (bottom row) is compared against various ablations. Bold and underline indicate the best and second-best performances.

We begin with a minimal setup that includes only the prediction head output ($\mathbf{P}^{\text{veh}}$, $\mathbf{P}^{\text{flow}}$) in the top row and progressively add other prediction tasks. As expected, vehicle prediction metrics score highest when the network is tasked solely with predicting these outputs. When DOGM predictions $\mathbf{Z}^{\text{ogm}}$ are included, a drop in vehicle prediction accuracy is observed, indicating a trade-off introduced by the added complexity. Notably, the inclusion of the detection head $\mathbf{D}_{\text{det}}$, together with generation of flow-guided occupancy grid predictions  $\mathbf{W}$ leads to substantial improvements in dynamic recall across multiple agent types, including vehicles, pedestrians, and cyclists. These components help mitigate the decline in vehicle prediction metrics seen with DOGM predictions. Meanwhile, the flow prediction error (EPE) remains stable across configurations, and the quality of dynamic occupancy predictions improves as more tasks are added.

Overall, the ablation study highlights the effectiveness of incorporating auxiliary tasks within a multi-head architecture. While additional tasks may introduce minor trade-offs in vehicle prediction accuracy, they significantly improve the model's ability to capture diverse dynamic elements in the scene, as we can see in the recall scores of dynamic agents in the warped grids. The overall architecture benefits from richer supervision, leading to more robust and generalizable motion and occupancy predictions.
\section{CONCLUSION} \label{sec:conclusion}
In an effort to bring together specialized and generic agent motion, a multi-task framework is built that leverages DOGMs and vehicle grids to predict the scene evolution. Future DOGMs, vehicle grids, and scene flow grids provide complementary and redundant scene information, collectively enhancing predictions by capturing patterns and behaviors based on both the observable static scene and motion of dynamic agents. Our approach effectively bridges the gap between agent-specific and generic motion prediction, demonstrating that holistic scene understanding is effectively captured by an interdependent loss formulation that enforces multi-task consistency. Evaluation on real-world datasets shows that flow-guided occupancy predictions improve the network's ability to predict vehicle as well as unrecognized agent motion compared to baseline video prediction methods. The results demonstrate that our method generalizes well across different datasets, consistently showing strong performance in dynamic motion prediction.

Future work will explore the inclusion of additional agent semantics in both the input and predicted grids, such as pedestrians and cyclists. Integrating online road map information could further enhance agent-specific predictions. Finally, predicting a flow grid that yields strictly kinematically feasible motion vectors for agents, along with the development of evaluation metrics that assess performance on the basis of such kinematic feasibility, is an interesting research direction to explore.

 

\bibliographystyle{IEEEtran}
\bibliography{biblio/jrnl}%

\begin{thebibliography}{10}
\providecommand{\url}[1]{#1}
\csname url@samestyle\endcsname
\providecommand{\newblock}{\relax}
\providecommand{\bibinfo}[2]{#2}
\providecommand{\BIBentrySTDinterwordspacing}{\spaceskip=0pt\relax}
\providecommand{\BIBentryALTinterwordstretchfactor}{4}
\providecommand{\BIBentryALTinterwordspacing}{\spaceskip=\fontdimen2\font plus
\BIBentryALTinterwordstretchfactor\fontdimen3\font minus \fontdimen4\font\relax}
\providecommand{\BIBforeignlanguage}[2]{{%
\expandafter\ifx\csname l@#1\endcsname\relax
\typeout{** WARNING: IEEEtran.bst: No hyphenation pattern has been}%
\typeout{** loaded for the language `#1'. Using the pattern for}%
\typeout{** the default language instead.}%
\else
\language=\csname l@#1\endcsname
\fi
#2}}
\providecommand{\BIBdecl}{\relax}
\BIBdecl

\bibitem{yurtsever2020survey}
E.~Yurtsever, J.~Lambert, A.~Carballo, and K.~Takeda, ``A survey of autonomous driving: Common practices and emerging technologies,'' \emph{IEEE access}, vol.~8, pp. 58\,443--58\,469, 2020.

\bibitem{mozaffari2020deep}
S.~Mozaffari, O.~Y. Al-Jarrah, M.~Dianati, P.~Jennings, and A.~Mouzakitis, ``Deep learning-based vehicle behavior prediction for autonomous driving applications: A review,'' \emph{IEEE Trans. Intelligent Transportation Systems}, vol.~23, no.~1, pp. 33--47, 2020.

\bibitem{lukas22}
L.~Rummelhard, J.-A. David, A.~G. Moreno, and C.~Laugier, ``A cross-prediction, hidden-state-augmented approach for dynamic occupancy grid filtering,'' in \emph{Proc. IEEE Int. Conf. on Control, Automation, Robotics and Vision (ICARCV)}, 2022, pp. 41--46.

\bibitem{asghar2022allo}
R.~Asghar, L.~Rummelhard, A.~Spalanzani, and C.~Laugier, ``Allo-centric occupancy grid prediction for urban traffic scene using video prediction networks,'' in \emph{Proc. IEEE Int. Conf. on Control, Automation, Robotics and Vision (ICARCV)}.\hskip 1em plus 0.5em minus 0.4em\relax IEEE, 2022, pp. 255--260.

\bibitem{asghar2023vehicle}
R.~Asghar, M.~Diaz-Zapata, L.~Rummelhard, A.~Spalanzani, and C.~Laugier, ``Vehicle motion forecasting using prior information and semantic-assisted occupancy grid maps,'' in \emph{Proc. IEEE/RSJ Int. Conf. on Intelligent Robots and Systems (IROS)}.\hskip 1em plus 0.5em minus 0.4em\relax IEEE, 2023, pp. 49--54.

\bibitem{toyungyernsub2024predicting}
M.~Toyungyernsub, E.~Yel, J.~Li, and M.~J. Kochenderfer, ``Predicting future spatiotemporal occupancy grids with semantics for autonomous driving,'' in \emph{Proc. IEEE Intelligent Vehicles Symposium (IV)}.\hskip 1em plus 0.5em minus 0.4em\relax IEEE, 2024, pp. 2855--2861.

\bibitem{jeon2018traffic}
H.-S. Jeon, D.-S. Kum, and W.-Y. Jeong, ``Traffic scene prediction via deep learning: Introduction of multi-channel occupancy grid map as a scene representation,'' in \emph{Proc. IEEE Intelligent Vehicles Symposium (IV)}.\hskip 1em plus 0.5em minus 0.4em\relax IEEE, 2018, pp. 1496--1501.

\bibitem{dequaire2017deep}
J.~Dequaire, P.~Ondr{\'u}{\v{s}}ka, D.~Rao, D.~Wang, and I.~Posner, ``Deep tracking in the wild: End-to-end tracking using recurrent neural networks,'' \emph{The International Journal of Robotics Research}, vol.~37, no. 4-5, pp. 492--512, 2017.

\bibitem{mohajerin2019multi}
N.~Mohajerin and M.~Rohani, ``Multi-step prediction of occupancy grid maps with recurrent neural networks,'' in \emph{Proc. IEEE/CVF Conf. on Computer Vision and Pattern Recognition (CVPR)}, 2019, pp. 10\,600--10\,608.

\bibitem{itkina2019dynamic}
M.~Itkina, K.~Driggs-Campbell, and M.~J. Kochenderfer, ``Dynamic environment prediction in urban scenes using recurrent representation learning,'' in \emph{Proc. IEEE Int. Conf. on Intelligent Transportation Systems (ITSC)}.\hskip 1em plus 0.5em minus 0.4em\relax IEEE, 2019, pp. 2052--2059.

\bibitem{reid2019localization}
T.~G. Reid, S.~E. Houts, R.~Cammarata, G.~Mills, S.~Agarwal, A.~Vora, and G.~Pandey, ``Localization requirements for autonomous vehicles,'' \emph{arXiv:1906.01061}, 2019.

\bibitem{Toyungyernsub2022}
M.~Toyungyernsub, E.~Yel, J.~Li, and M.~J. Kochenderfer, ``Dynamics-aware spatiotemporal occupancy prediction in urban environments,'' in \emph{Proc. IEEE/RSJ Int. Conf. on Intelligent Robots and Systems (IROS)}, 2022, pp. 10\,836--10\,841.

\bibitem{schreiber2019long}
M.~Schreiber, S.~Hoermann, and K.~Dietmayer, ``Long-term occupancy grid prediction using recurrent neural networks,'' in \emph{Proc. IEEE/RSJ Int. Conf. on Robotics and Automation (ICRA)}.\hskip 1em plus 0.5em minus 0.4em\relax IEEE, 2019, pp. 9299--9305.

\bibitem{lange2021attention}
B.~Lange, M.~Itkina, and M.~J. Kochenderfer, ``Attention augmented convlstm for environment prediction,'' in \emph{Proc. IEEE/RSJ Int. Conf. on Intelligent Robots and Systems (IROS)}.\hskip 1em plus 0.5em minus 0.4em\relax IEEE, 2021, pp. 1346--1353.

\bibitem{gao2020vectornet}
J.~Gao, C.~Sun, H.~Zhao, Y.~Shen, D.~Anguelov, C.~Li, and C.~Schmid, ``Vectornet: Encoding hd maps and agent dynamics from vectorized representation,'' in \emph{Proc. IEEE/CVF Conf. on Computer Vision and Pattern Recognition (CVPR)}, 2020, pp. 11\,525--11\,533.

\bibitem{salzmann2020trajectron++}
T.~Salzmann, B.~Ivanovic, P.~Chakravarty, and M.~Pavone, ``Trajectron++: Dynamically-feasible trajectory forecasting with heterogeneous data,'' in \emph{Proc. Eur. Conf. on Computer Vision (ECCV)}.\hskip 1em plus 0.5em minus 0.4em\relax Springer, 2020, pp. 683--700.

\bibitem{varadarajan2022multipath++}
B.~Varadarajan, A.~Hefny, A.~Srivastava, K.~S. Refaat, N.~Nayakanti, A.~Cornman, K.~Chen, B.~Douillard, C.~P. Lam, D.~Anguelov \emph{et~al.}, ``Multipath++: Efficient information fusion and trajectory aggregation for behavior prediction,'' in \emph{Proc. IEEE/RSJ Int. Conf. on Robotics and Automation (ICRA)}.\hskip 1em plus 0.5em minus 0.4em\relax IEEE, 2022, pp. 7814--7821.

\bibitem{schafer2022context}
M.~Sch{\"a}fer, K.~Zhao, M.~B{\"u}hren, and A.~Kummert, ``Context-aware scene prediction network (caspnet),'' in \emph{Proc. IEEE Int. Conf. on Intelligent Transportation Systems (ITSC)}.\hskip 1em plus 0.5em minus 0.4em\relax IEEE, 2022, pp. 3970--3977.

\bibitem{diaz2022hd}
A.~Diaz-Diaz, M.~Ocana, A.~Llamazares, C.~G{\'o}mez-Hu{\'e}lamo, P.~Revenga, and L.~M. Bergasa, ``Hd maps: Exploiting opendrive potential for path planning and map monitoring,'' in \emph{Proc. IEEE Intelligent Vehicles Symposium (IV)}.\hskip 1em plus 0.5em minus 0.4em\relax IEEE, 2022, pp. 1211--1217.

\bibitem{schmidt2022crat}
J.~Schmidt, J.~Jordan, F.~Gritschneder, and K.~Dietmayer, ``Crat-pred: Vehicle trajectory prediction with crystal graph convolutional neural networks and multi-head self-attention,'' in \emph{Proc. IEEE/RSJ Int. Conf. on Robotics and Automation (ICRA)}.\hskip 1em plus 0.5em minus 0.4em\relax IEEE, 2022, pp. 7799--7805.

\bibitem{benrachou2023improving}
D.~E. Benrachou, S.~Glaser, M.~Elhenawy, and A.~Rakotonirainy, ``Improving efficiency and generalisability of motion predictions with deep multi-agent learning and multi-head attention,'' \emph{IEEE Trans. Intelligent Transportation Systems}, vol.~25, no.~6, pp. 5356--5373, 2023.

\bibitem{gomez2023improving}
C.~G{\'o}mez-Hu{\'e}lamo, M.~V. Conde, R.~Barea, and L.~M. Bergasa, ``Improving multi-agent motion prediction with heuristic goals and motion refinement,'' in \emph{Proc. IEEE/CVF Conf. on Computer Vision and Pattern Recognition (CVPR)}, 2023, pp. 5323--5332.

\bibitem{gomez2023efficient}
C.~G{\'o}mez-Hu{\'e}lamo, M.~V. Conde, R.~Guti{\'e}rrez-Moreno, R.~Barea, A.~Llamazares, M.~Antunes, and L.~M. Bergasa, ``Efficient context-aware graph transformer for vehicle motion prediction,'' in \emph{Proc. IEEE Int. Conf. on Intelligent Transportation Systems (ITSC)}.\hskip 1em plus 0.5em minus 0.4em\relax IEEE, 2023, pp. 4133--4140.

\bibitem{liang2020pnpnet}
M.~Liang, B.~Yang, W.~Zeng, Y.~Chen, R.~Hu, S.~Casas, and R.~Urtasun, ``Pnpnet: End-to-end perception and prediction with tracking in the loop,'' in \emph{Proc. IEEE/CVF Conf. on Computer Vision and Pattern Recognition (CVPR)}, 2020, pp. 11\,553--11\,562.

\bibitem{wu2020motionnet}
P.~Wu, S.~Chen, and D.~N. Metaxas, ``Motionnet: Joint perception and motion prediction for autonomous driving based on bird's eye view maps,'' in \emph{Proc. IEEE/CVF Conf. on Computer Vision and Pattern Recognition (CVPR)}, 2020, pp. 11\,385--11\,395.

\bibitem{luo2018fast}
W.~Luo, B.~Yang, and R.~Urtasun, ``Fast and furious: Real time end-to-end 3d detection, tracking and motion forecasting with a single convolutional net,'' in \emph{Proc. IEEE/CVF Conf. on Computer Vision and Pattern Recognition (CVPR)}, 2018, pp. 3569--3577.

\bibitem{fiery2021}
A.~Hu, Z.~Murez, N.~Mohan, S.~Dudas, J.~Hawke, V.~Badrinarayanan, R.~Cipolla, and A.~Kendall, ``{FIERY}: Future instance segmentation in bird's-eye view from surround monocular cameras,'' in \emph{Proceedings of the International Conference on Computer Vision ({ICCV})}, 2021.

\bibitem{casas2021mp3}
S.~Casas, A.~Sadat, and R.~Urtasun, ``Mp3: A unified model to map, perceive, predict and plan,'' in \emph{Proc. IEEE/CVF Conf. on Computer Vision and Pattern Recognition (CVPR)}, 2021, pp. 14\,403--14\,412.

\bibitem{hendy2020fishing}
N.~Hendy, C.~Sloan, F.~Tian, P.~Duan, N.~Charchut, Y.~Xie, C.~Wang, and J.~Philbin, ``Fishing net: Future inference of semantic heatmaps in grids,'' \emph{arXiv:2006.09917}, 2020.

\bibitem{mahjourian2022occupancy}
R.~Mahjourian, J.~Kim, Y.~Chai, M.~Tan, B.~Sapp, and D.~Anguelov, ``Occupancy flow fields for motion forecasting in autonomous driving,'' \emph{IEEE Robotics and Automation Letters}, vol.~7, no.~2, pp. 5639--5646, 2022.

\bibitem{mann2022predicting}
K.~S. Mann, A.~Tomy, A.~Paigwar, A.~Renzaglia, and C.~Laugier, ``Predicting future occupancy grids in dynamic environment with spatio-temporal learning,'' in \emph{Proc. IEEE Intelligent Vehicles Symposium (IV)}.\hskip 1em plus 0.5em minus 0.4em\relax IEEE, 2022, pp. 1121--1126.

\bibitem{asghar2024flow}
R.~Asghar, W.~Liu, L.~Rummelhard, A.~Spalanzani, and C.~Laugier, ``Flow-guided motion prediction with semantics and dynamic occupancy grid maps,'' in \emph{Proc. IEEE Int. Conf. on Intelligent Transportation Systems (ITSC)}, 2024, pp. 3160--3165.

\bibitem{philion2020lift}
J.~Philion and S.~Fidler, ``Lift, splat, shoot: Encoding images from arbitrary camera rigs by implicitly unprojecting to 3d,'' in \emph{Proc. Eur. Conf. on Computer Vision (ECCV)}.\hskip 1em plus 0.5em minus 0.4em\relax Springer, 2020, pp. 194--210.

\bibitem{lee2021video}
S.~Lee, H.~G. Kim, D.~H. Choi, H.-I. Kim, and Y.~M. Ro, ``Video prediction recalling long-term motion context via memory alignment learning,'' in \emph{Proc. IEEE/CVF Conf. on Computer Vision and Pattern Recognition (CVPR)}, 2021, pp. 3054--3063.

\bibitem{hu2020probabilistic}
A.~Hu, F.~Cotter, N.~Mohan, C.~Gurau, and A.~Kendall, ``Probabilistic future prediction for video scene understanding,'' in \emph{Proc. Eur. Conf. on Computer Vision (ECCV)}.\hskip 1em plus 0.5em minus 0.4em\relax Springer, 2020, pp. 767--785.

\bibitem{ballas2015delving}
N.~Ballas, L.~Yao, C.~Pal, and A.~Courville, ``Delving deeper into convolutional networks for learning video representations,'' \emph{arXiv:1511.06432}, 2015.

\bibitem{nuscenes2019}
H.~Caesar, V.~Bankiti, A.~H. Lang, S.~Vora, V.~E. Liong, Q.~Xu, A.~Krishnan, Y.~Pan, G.~Baldan, and O.~Beijbom, ``nuscenes: A multimodal dataset for autonomous driving,'' in \emph{Proc. IEEE/CVF Conf. on Computer Vision and Pattern Recognition (CVPR)}, 2020, pp. 11\,621--11\,631.

\bibitem{WovenPlanet2019}
R.~Kesten, M.~Usman, J.~Houston, T.~Pandya, K.~Nadhamuni, A.~Ferreira, M.~Yuan, B.~Low, A.~Jain, P.~Ondruska, S.~Omari, S.~Shah, A.~Kulkarni, A.~Kazakova, C.~Tao, L.~Platinsky, W.~Jiang, and V.~Shet, ``Woven planet perception dataset 2020,'' \url{https://woven.toyota/en/perception-dataset}, 2019.

\bibitem{lotter2016deep}
W.~Lotter, G.~Kreiman, and D.~Cox, ``Deep predictive coding networks for video prediction and unsupervised learning,'' \emph{arXiv:1605.08104}, 2016.

\bibitem{wang2022predrnn}
Y.~Wang, H.~Wu, J.~Zhang, Z.~Gao, J.~Wang, S.~Y. Philip, and M.~Long, ``Predrnn: A recurrent neural network for spatiotemporal predictive learning,'' \emph{IEEE Trans. Pattern Analysis and Machine Intelligence}, vol.~45, no.~2, pp. 2208--2225, 2022.

\bibitem{girase2021physically}
H.~Girase, J.~Hoang, S.~Yalamanchi, and M.~Marchetti-Bowick, ``Physically feasible vehicle trajectory prediction,'' \emph{arXiv:2104.14679}, 2021.

\end{thebibliography}

\vfill

\end{document}